\documentclass[journal]{IEEEtran}
\usepackage{cite}
\usepackage{amsmath, amsfonts}
\usepackage{algorithmic}
\usepackage{graphicx}
\usepackage{textcomp}
\usepackage{xcolor}
\usepackage[caption=false,font=normalsize,labelfont=sf,textfont=sf]{subfig}

\usepackage{tikz}
\usepackage{url}
\usepackage{hyperref} 
\hypersetup{hidelinks,
    colorlinks=true,
    allcolors=black,
    pdfstartview=Fit,
    breaklinks=true
}

\usepackage{siunitx}
\usepackage{multirow}
\usepackage{makecell}
\usepackage{booktabs}
\usepackage{float}
\usepackage{gensymb}

\usepackage[linesnumbered,ruled,vlined]{algorithm2e}
\usepackage{graphicx}
\usepackage{enumitem}

\SetCommentSty{emph} 
\newcommand{\mycomment}[1]{\textcolor{green!50!black}{#1}} 

\begin{document}
\title{
Freehand 3D Ultrasound Imaging: Sim-in-the-Loop Probe Pose Optimization via Visual Servoing
}
\author{Yameng Zhang, Dianye Huang, Max Q.-H. Meng, \IEEEmembership{Fellow, IEEE}, \\ Nassir Navab, \IEEEmembership{Fellow, IEEE}, and Zhongliang Jiang
\thanks{This study was partly supported by the Multiscale Medical Robotics Centre, AIR@InnoHK and SINO-German Mobility Project under Grant M0221.}
\thanks{Yameng Zhang is with the Department of Mechanical Engineering, The University of Hong Kong, Hong Kong SAR, China, and also with the Department of Electronic Engineering, The Chinese University of Hong Kong (CUHK), Hong Kong SAR, China
(e-mail: zhangyameng@link.cuhk.edu.hk).}
\thanks{Dianye Huang and Nassir Navab are with the Chair for Computer Aided Medical Procedures and Augmented Reality (CAMP), Technical University of Munich (TUM), Munich 80333, Germany (e-mail: dianye.huang@tum.de; nassir.navab@tum.de).}
\thanks{Max Q.-H. Meng is with the Department of Electronic and Electrical Engineering, Southern University of Science and Technology, Shenzhen 518055, China, and also with the Department of Electronic Engineering, CUHK, Hong Kong SAR, China (e-mail: max.meng@ieee.org).}
\thanks{Zhongliang Jiang is with the Department of Mechanical Engineering, The University of Hong Kong, Hong Kong SAR, China (corresponding author, e-mail: zljiang@hku.hk).}
\thanks{Ethical approval of the robotic scans on limbs was granted by the Institutional Review Board from the TUM under Application No. 2022-87-S-KK, and performed in line with the Declaration of Helsinki.}
}

\maketitle

\begin{abstract}
Freehand 3D ultrasound (US) imaging using conventional 2D probes offers flexibility and accessibility for diverse clinical applications but faces challenges in accurate probe pose estimation. Traditional methods depend on costly tracking systems, while neural network-based methods struggle with image noise and error accumulation, compromising reconstruction precision. We propose a cost-effective and versatile solution that leverages lightweight cameras and visual servoing in simulated environments for precise 3D US imaging. These cameras capture visual feedback from a textured planar workspace. To counter occlusions and lighting issues, we introduce an image restoration method that reconstructs occluded regions by matching surrounding texture patterns. For pose estimation, we develop a simulation-in-the-loop approach, which replicates the system setup in simulation and iteratively minimizes pose errors between simulated and real-world observations. A visual servoing controller refines the alignment of camera views, improving translational estimation by optimizing image alignment. Validations on a soft vascular phantom, a 3D-printed conical model, and a human arm demonstrate the robustness and accuracy of our approach, with Hausdorff distances to the reference reconstructions of 0.359 mm, 1.171 mm, and 0.858 mm, respectively. These results confirm the method's potential for reliable freehand 3D US reconstruction. Project resources are available at \url{https://github.com/YamengZZZ/Fh3DUS}.
\end{abstract}


\section{Introduction}
\begin{figure}[t]
\centering
\includegraphics[width=0.48\textwidth]{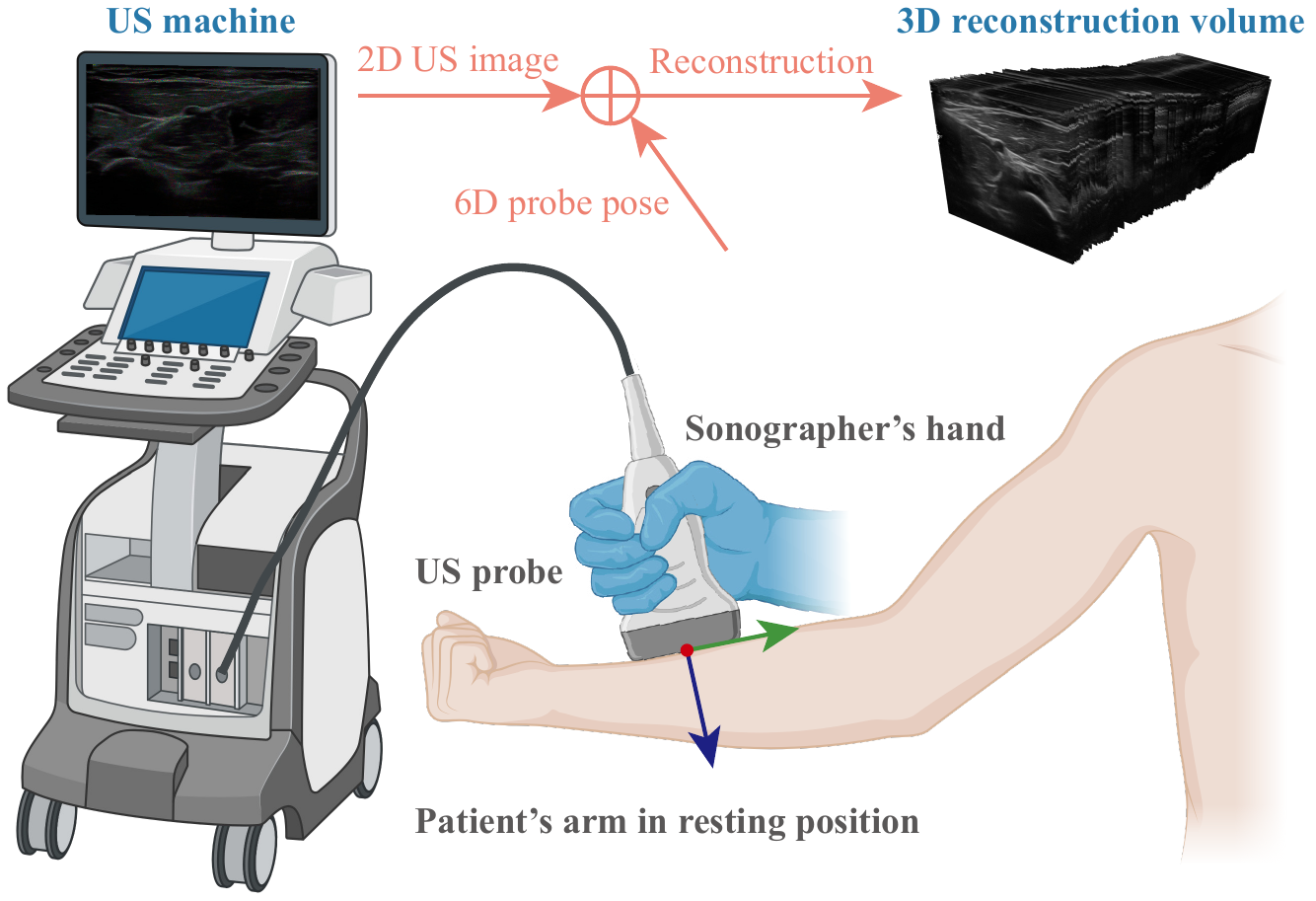}
\caption{Illustration of a freehand 3D US imaging system using a conventional 2D US probe. The acquired 2D images can be stacked to reconstruct 3D volumes based on positional tracking data from tracking devices or pose estimation algorithms.
}
\label{fig:freehand}
\vspace{-1.0em}
\end{figure}

\IEEEPARstart{M}{edical} ultrasound (US) is widely used in modern clinical practice due to its low cost, real-time imaging, and lack of ionizing radiation. It serves as a first-line tool in various applications, including obstetrics and emergency medicine. Given its versatility, US has been seen as a promising imaging solution to address the global shortage of primary healthcare interventions, particularly in low-income regions~\cite{jiang2023robotic}. However, the interpretation of 2D US images is often challenging due to the unavoidable speckle artifacts and echo interference. To provide intuitive anatomical details, 3D US imaging has gained increasing attention in the community. A representative free-hand 3D US imaging scenario is depicted in Fig. \ref{fig:freehand}.

Methods for generating 3D US volumes can be broadly classified into two approaches: matrix array probes, which provide native 3D volumes, and tracking-based 3D imaging. Research on matrix array probes dates back to the 1990s when 2D array transducers were developed for volumetric imaging~\cite{hossack2002quantitative}. {Recent advances in materials science have enabled the development of innovative devices. For example, Hu~\emph{et al.} developed a soft wearable US patch for 3D cardiac imaging~\cite{hu2023wearable}. However, due to high cost, time inefficiency, complex signal interference, and limited field of view (FoV), native 3D probes have not been widely adopted in clinical practice.}

As a practical alternative, tracking-based 3D imaging is often regarded as a promising solution. A commonly used approach is the wobbler probe, which generates 3D images based on mechanical tracking data~\cite{chatelain2017confidence}. However, since the 1D element must be rotated or tilted, this method suffers from a limited frame rate. To address this limitation and maintain the flexibility of US scanning, external tracking systems, such as electromagnetic (EM)~\cite{lang2009fusion, shangyi2024tmech} and optical devices~\cite{guerrero2007real, chenguang2025tase}, have been introduced to enable freehand 3D US acquisition. The obtained 2D images are stacked based on their paired tracking data to reconstruct volumetric data in 3D space~\cite{lasso2014plus}. Nevertheless, these methods rely on costly tracking devices. Additionally, to have consistent and accurate tracking data, users may need to adapt their workflow to avoid optical occlusion and electromagnetic interference. With recent advances in robotics, Jiang~\emph{et al.} adapted a robotic manipulator to compute 3D volumetric data based on 2D images~\cite{jiang2024intelligent}. Similar robotic setups have been developed for various clinical applications~\cite{huang2018robotic,chen2023fully, jiang2022towards, ma2024guiding, huang2024robot, tan2023autonomous, jiang2021deformation, jiang2023defcor, dyck2024toward, yang2025tmech}. Despite these advancements, the increased cost requirement to work together with robots restrict its wide deployment in hospitals or emergency units.

To keep the US merits of operational flexibility and low cost in 3D imaging, image-based freehand 3D US reconstruction has been investigated recently. Guo~\emph{et al.} proposed a learning-based framework DCL-Net to predict the transformation between two frames~\cite{guo2020sensorless}. {This method aims to achieve precise pose estimation based solely on image inputs. Although it outperforms the baseline methods, the average distance error of approximately $10.33~\mathrm{mm}$ is not suitable for accurate 3D reconstruction.} To enhance the pose estimation accuracy, Prevost~\emph{et al.} combined the 2D US images and an Inertial Measurement Unit (IMU) into a neural network to estimate the motion between successive US frames~\cite{prevost20183d}. To explicitly account for the scanning pattern, Luo~\emph{et al.} proposed an online learning framework that recurrently optimizes the final reconstruction results~\cite{luo2021self}. To investigate the long-term dependency for sensorless 3D reconstruction, Li~\emph{et al.} proposed a sequence model with multi-transformation prediction~\cite{li2023long}. Despite the achievements, these methods are not ready for use because the pose estimation in each pair of frames will be accumulated in the final reconstruction, leading to a severe reduction of final 3D image accuracy. To alleviate this issue, Luo~\emph{et al.} used multiple IMU measurements as weak labels for adaptive pose optimization~\cite{luo2022deep}. While this framework significantly reduced reconstruction drift, the reported translational and rotational errors of $10.24~\mathrm{mm}$ and $1.55^{\circ}$, respectively, remain inadequate for clinical use. In addition to using US images, Sun~\emph{et al.} employed an optical flow technique to estimate probe poses using camera images. However, due to accumulated estimation errors, the final translation error exceeded $9~\mathrm{mm}$ and the rotation error surpassed $5^{\circ}$ after a probe movement of just $100~\mathrm{mm}$~\cite{sun2014probe}. Recently, Huang~\emph{et al.} addressed this challenge by introducing PoseNet, which directly predicts the probe's 6D pose based on dual camera observations. However, its reliance on large training datasets limits practical applicability \cite{huang2025improving}.

To address the challenges, we propose a novel freehand 3D US reconstruction method with the following contributions:
\begin{enumerate}
    \item We introduce a cost-effective and versatile US probe localization system that integrates two lightweight cameras with the probe. {These cameras capture visual feedback from a planar workspace with a complex-textured RGB pattern, enabling precise and robust pose estimation.}
    \item We propose a simulation-in-the-loop (sim-in-the-loop) US probe localization method that treats localization as a visual feedback control problem. Real-world camera images serve as targets, while a position-based visual servoing (PBVS) controller iteratively adjusts the simulated probe pose to minimize discrepancies between simulated and real-world camera observations. The estimated probe pose is extracted from the simulation once the pose error drops below a predefined threshold.
    \item To address occlusions and lighting disturbances during object scanning, we develop an image restoration method that reconstructs occluded regions by matching surrounding texture patterns. Additionally, to correct pose errors arising from Sim2Real discrepancies, we introduce a calibration method that maps simulation-derived poses to ground-truth values.
    \item {Our method estimates the 6D pose of the US probe in a global reference frame, effectively eliminating the challenges of accumulated errors in recent image-based approaches~\cite{guo2020sensorless, prevost20183d, luo2023recon} during long sweeps in 3D reconstruction.}
\end{enumerate}

To demonstrate the practical applicability of the proposed method for freehand US scanning, we perform scans on three different objects: a soft vascular phantom, a 3D-printed conical model, and a human arm. The resulting Hausdorff distances to the reference reconstructions are 0.359 mm, 1.171 mm, and 0.858 mm, respectively. The intuitive demonstration of the freehand reconstruction results can be found in this video\footnote{\url{https://youtu.be/5md2VlLgI88}}.

The remainder of this paper is organized as follows: Section \ref{sec:system} presents an overview of our proposed system in both real-world and simulation settings and defines the key problems addressed. Section \ref{sec:method} details the US probe localization method. Section \ref{sec:exp} reports real-world experiments demonstrating the method’s precision and practicality. Section \ref{sec:discussion} discusses its advantages, limitations, and potential improvements. Finally, Section \ref{sec:conclusion} concludes the paper.

\section{System Overview and Problem Definition}
\label{sec:system}
This section introduces our US probe pose estimation system, ensuring consistent setups across both real-world and simulation environments. We also outline the motivation for integrating a simulation environment and highlight the key challenges our method seeks to overcome.

\begin{table*}[t]
\centering
\caption{{Recent Advances in Freehand 3D US Reconstruction}}
\label{tab:comp_err}
\resizebox{0.95\textwidth}{!}{
\begin{tabular}{l l l l}
\toprule
\textbf{Method} & \textbf{Year} & \textbf{Modality} & \textbf{Error Metrics} \\
\midrule
Skin feature tracking \cite{sun2014probe} & \textit{MICCAI} 2014 & Camera & Trans.: $9.1 \pm 4.9~\mathrm{mm}$, Rot.: $5.5^{\circ} \pm 1.7^{\circ}$ per $100~\mathrm{mm}$ \\
CNN-based motion estimation \cite{prevost20183d} & \textit{MedIA} 2018 & US + IMU & Trans.: $5.2~\mathrm{mm}$ per $100~\mathrm{mm}$, Rot.: N/A \\
Deep Contextual Learning Network (DCL-Net) \cite{guo2020sensorless} & \textit{MICCAI} 2020 & US & Trans.: $10.33~\mathrm{mm}$, Max Drift: $17.39~\mathrm{mm}$ \\
Deep Motion Network (MoNet) \cite{luo2022deep} & \textit{MICCAI} 2022 & US + IMU & Trans.: $10.24 \pm 7.36~\mathrm{mm}$, Rot.: $1.55^{\circ} \pm 1.46^{\circ}$ \\
Long-term sequence modeling \cite{li2023long} & \textit{TBME} 2023 & US & Trans.: $9.44 \pm 0.50~\mathrm{mm}$ (using 74 past frames) \\
Online learning framework (RecON) \cite{luo2023recon} & \textit{MedIA} 2023 & US & Trans.: $18.69 \pm 12.78~\mathrm{mm}$, Rot.: $2.26^{\circ} \pm 2.66^{\circ}$ \\
Vision and IMU fusion (UKF-LG) \cite{he2024freehand} & \textit{Ultra. Med.} 2023 & Camera + IMU & Trans.: $3.78~\mathrm{mm}$, Rot.: $0.36^{\circ}$ \\
Novel Coupling Pad \cite{dai2024advancing} & \textit{MICCAI} 2024 & US & Trans.: $2.52~\mathrm{mm}$, Rot.: N/A \\
Cross Encoder Decoder (CED) \cite{huang2025improving} & \textit{ICRA} 2025 & Camera & Trans.: $2.03 ~\mathrm{mm}$, Rot.: $0.37^{\circ}$ \\
\bottomrule
\end{tabular}
}
\end{table*}

\subsection{System Overview}
\label{subsec:system}
\begin{figure}[t]
\centering
\includegraphics[width=0.42\textwidth]{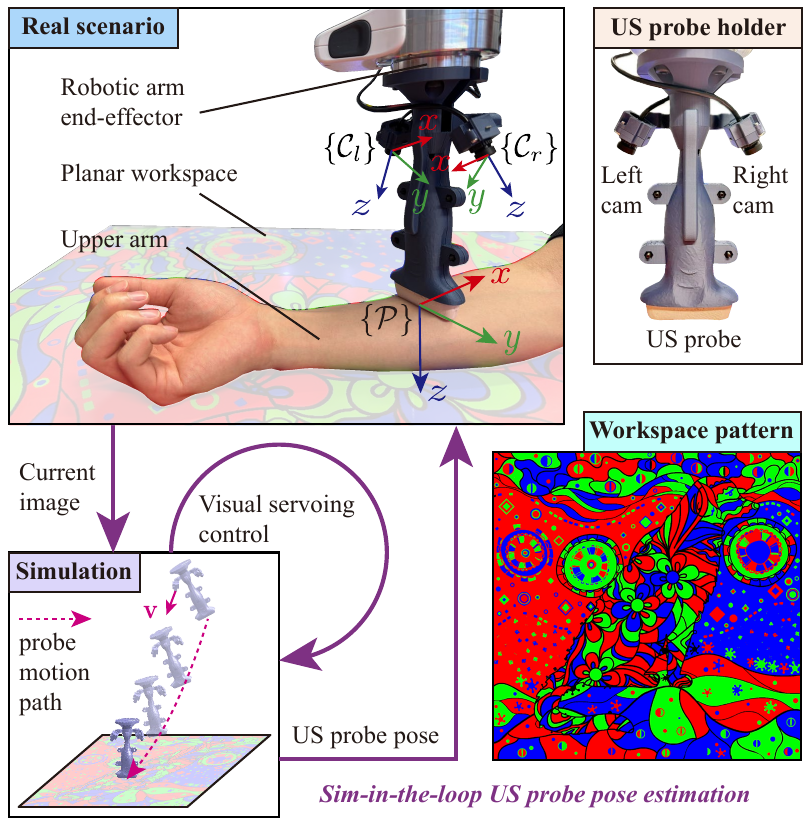}
\caption{Sim-in-the-loop US probe pose estimation system: Two monocular cameras on the probe capture visual features from a planar RGB-patterned workspace. Real-world images are transmitted to the simulation, where visual servoing aligns them with simulated images, {enabling extraction of the probe pose in the global frame.}}
\label{fig:system}
\vspace{-1.0em}
\end{figure}

As shown in Fig. \ref{fig:system}, the proposed US probe pose estimation system consists of two main components: a real-world scanning scenario and a corresponding simulated environment. In the real-world scenario, the system captures RGB images using {two cost-effective monocular cameras (2303U, China, \texteuro15 each),} each with a resolution of $360 \times 640$ pixels, a horizontal FoV of $83^{\circ}$, and a frame rate of 30 frames per second (fps). {The low-resolution, cost-effective cameras were chosen to balance system affordability, accuracy, and computational efficiency. While higher resolutions might improve accuracy, they would significantly increase processing time and cost. The selected resolution is sufficient for reliable localization with manageable computational demands.} These cameras are mounted symmetrically on a 3D-printed US probe holder in opposite directions to minimize visual obstructions from operators or patients. The US probe’s reference frame, denoted $\{\mathcal{P}\}$, has its $z$-axis aligned with the probe centerline and its $y$-axis along the front probe edge. {The left camera’s frame, $\{\mathcal{C}_l\}$, is oriented relative to $\{\mathcal{P}\}$ with a roll, pitch, and yaw of $(30^{\circ}, 10^{\circ}, 0)$, while the right camera’s frame, $\{\mathcal{C}_r\}$, has a relative pose of $(-30^{\circ}, 10^{\circ}, 180^{\circ})$. These angles were empirically chosen for optimal observability and reduced occlusion.} To accurately model the camera system, both intrinsic and eye-in-hand calibrations are performed to determine the camera parameters and the transformation from the probe frame to each camera frame. These calibration results are then used to configure the virtual camera settings in the simulation.

To enhance visual distinctiveness for camera observations, the planar workspace features a textured surface with intricate RGB patterns (see Fig.~\ref{fig:system}), designed to ensure distinguishable visual observations at individual viewpoints. The base pattern was obtained from \href{https://create.vista.com/unlimited/stock-vectors/247139728/stock-vector-hand-drawn-intricate-texture-abstract-patterns-asian-background-dog-isolated/}{create.vista.com}, then vectorized and randomly filled with RGB colors to further enhance feature diversity. In this study, the workspace measures $800 \times 800~\mathrm{mm}$. For individual applications, both the texture and dimensions of this planar workspace can be further customized as needed. 

The US transducer used in this study is a linear array transducer (12L3, ACUSON Juniper, SIEMENS AG, Germany). For ground-truth acquisition during scans, the transducer is mounted on a robotic arm (Franka Emika Panda, Franka GmbH, Germany) using a 3D-printed probe holder. {Notably, the robotic arm is used exclusively for quantitative evaluation and calibration purposes. It is not used in probe pose estimation, meaning that the proposed method is applicable in a hand-held scanning scenario.} US images are acquired via a frame grabber (MAGEWELL, China) and streamed at 30 fps. These images are not used for probe pose estimation, but are instead utilized to visualize the reconstructed 3D volumes based on the estimated probe poses. Alternative tracking devices could also serve as ground-truth providers for validation, highlighting the flexibility of the proposed evaluation framework. The precise robotic tracking, in combination with paired US images, is used to quantitatively and qualitatively validate the effectiveness of the proposed freehand 3D reconstruction method.

The simulation environment is built using CoppeliaSim\footnote{https://www.coppeliarobotics.com/}, a versatile and powerful robotic simulator. It replicates the real-world setup, including the intrinsic parameters of two cameras and their relative location on the probe holder. To maintain consistent visual features between the virtual and real environments, the simulator applies the same textured workspace pattern as the real scenario. For a unified coordinate description across both simulation and real-world cases, a global reference is established at the center of the workspace pattern.
 
Communication between the real and simulated environments is realized via the ROS interface\footnote{https://www.ros.org/}. Real-world camera images are published to ROS, while the simulator thread subscribes to this topic, using it as the target image. The virtual probe location is then iteratively adjusted within the simulator to find the optimal pose that replicates the real camera observation. This pose is then considered the final estimated probe pose in the global frame. Notably, the proposed probe pose estimation method is compatible with single-camera setups. In this study, two cameras are used to demonstrate the method's scalability and its improved performance in handling severe occlusions from specific directions.

\subsection{Practical Problem Definition}
\label{subsec:task}
Accurate probe pose estimation for freehand 3D US reconstruction remains a persistent challenge~\cite{sun2014probe, weiyong2025tase}. Recent deep learning approaches using B-mode images struggle with precision limitations~\cite{guo2020sensorless, prevost20183d, luo2021self, li2023long}. While in-plane motion can be tracked via pixel-level changes in consecutive images, out-of-plane pose estimation is hindered by insufficient visual cues. Moreover, relative pose estimation between frames inevitably accumulates errors, degrading the final 3D reconstruction \cite{dai2024advancing}. {A summary of recent advances in freehand 3D US reconstruction and their reported errors is provided in Table~\ref{tab:comp_err}. The relatively large errors hinder the practical use of these methods in real scenarios.}

To further advance the field, we propose a reliable method that uses low-cost external cameras and a simulated environment to estimate probe pose in the global reference frame. The method is designed to be versatile and does not currently require training on specific anatomical data for practical use. By leveraging simulation, we reframe pose estimation as a control task: given a real-world target image, the probe pose in simulation is iteratively optimized until the simulated camera observation aligns with the real image. This optimization is achieved using a visual servoing controller~\cite{chaumette2006visual, zhang2023servo, zhiwen2024tmech, hesheng2024tmech, giuseppe2021tmech, Zakeri2025tmech}, which minimizes the discrepancy between simulated and real images in a closed-loop manner. The proposed approach offers two key advantages. First, the simulator provides direct access to system parameters, such as camera-to-workspace distance at each control step, which are difficult to measure in real-world settings. Second, the closed-loop pose estimation framework continuously refines alignment between simulated and real images, enabling high accuracy with full spatial freedom. To ensure the method's translational capability, this study explicitly considers the following challenges:

\begin{enumerate}
    \item \textbf{Global pose estimation from a single observation}: Estimating the global probe pose using only the current camera observation is challenging due to the complex and ambiguous relationship between the image and the probe’s spatial configuration.

    \item \textbf{Variations in camera observations caused by occlusions and lighting conditions}: The patient or sonographer may partially or fully obstruct the camera's view, leading to inconsistencies in pose estimation. Additionally, variations in lighting conditions can significantly alter the visual appearance of camera images.
    
    \item \textbf{Sim2Real gap}: This gap arises from discrepancies in workspace configuration, as well as inaccuracies in camera intrinsics and hand-eye calibration, both of which affect pose estimation accuracy.
\end{enumerate}

\section{Probe Pose Estimation}
\label{sec:method}
\begin{figure*}[htp!]
\centering
\includegraphics[width=0.9\textwidth]{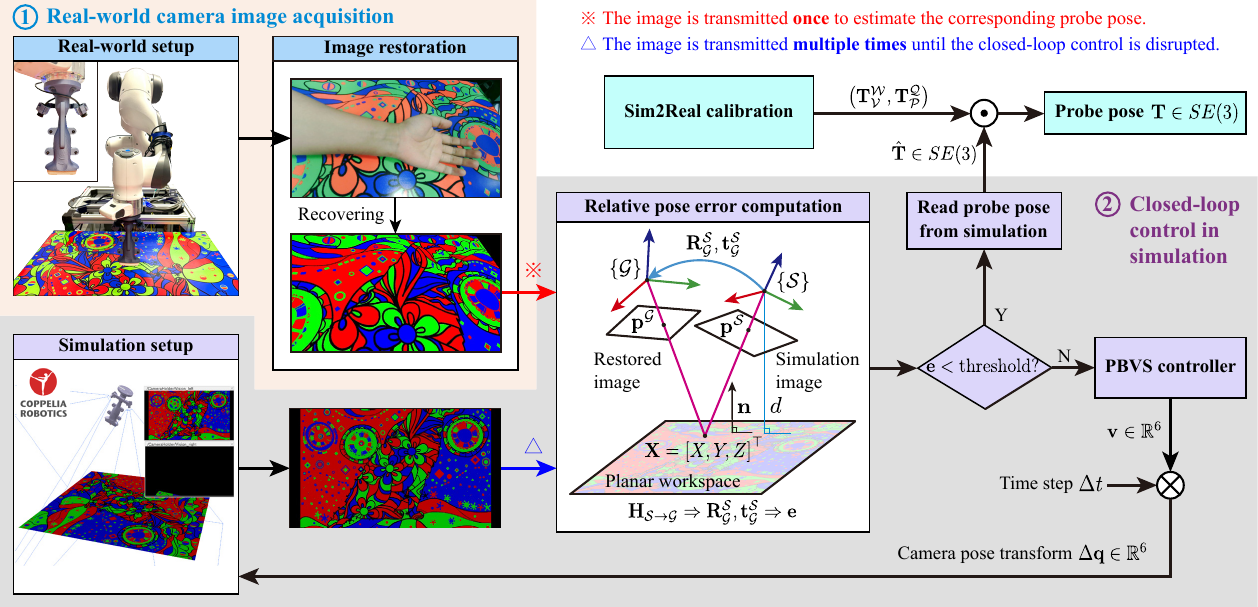}
\caption{Pipeline of the Probe Pose Estimation Algorithm. The process begins with image restoration, which recovers details obscured by objects or lighting reflections. Next, the pose error is computed by comparing the restored image with a simulated counterpart. A PBVS approach iteratively updates the probe's location in the simulation to achieve the best match with the restored real-world camera observation. By applying a one-time Sim2Real compensation to the optimized probe pose in simulation, we obtain a refined pose estimation. {The robotic arm is included solely for quantitative evaluation, not as part of the proposed pose estimation method.}}
\label{fig:pipeline}
\vspace{-1.0em}
\end{figure*}

The overall pipeline of the proposed low-cost, camera-based probe pose estimation is illustrated in Fig.~\ref{fig:pipeline}. Using prior knowledge of the complete workspace pattern, we generate an idealized restoration by warping the pattern to match the observed camera view, assuming a flat and undistorted workspace. This process yields a reference image for pose estimation that compensates for observed occlusions. The restored image then serves as the target for probe pose optimization, where the goal is to find an ideal probe pose that produces a simulated image matching the target. To achieve this, the probe is randomly initialized in simulation, and the relative pose error between the current simulated probe pose and the unknown ideal pose is computed using the restored and simulated images. A PBVS control algorithm iteratively refines the simulated pose to minimize this error. Once the error drops below a predefined threshold, the simulated pose is considered an approximation of the probe's global pose. The key components of the proposed method are detailed below.

\subsection{Recovering Standardized Camera Observation}
\label{subsec:restoration}
{To mitigate performance degradation from visual occlusions and lighting variations, we first generate an idealized restoration of the workspace using the known pattern and geometric registration. By utilizing real-time camera observations of the feature-rich textured workspace, we compute the homography matrix mapping the known workspace pattern to the current camera view. Applying this transformation to the full workspace pattern yields a standardized, idealized reference image that is used for pose estimation.}

The homography computation involves feature detection, description, matching, and parameter calculation. State-of-the-art (SOTA) feature detectors and descriptors can be categorized as handcrafted feature-based and learning-based. Although learning-based methods excel at finding correspondences in textureless areas and repetitive patterns, they demand extensive and domain-specific training. Handcrafted feature-based methods, conversely, are simpler to implement and require no data collection or training, making them more suitable for our application.

Among handcrafted feature-based methods, Accelerated KAZE (AKAZE)~\cite{akaze} provides a strong balance between speed and performance. It constructs a nonlinear scale space using nonlinear diffusion, preserving image details while handling noise. Keypoints are detected using the Hessian matrix, and a gradient-based descriptor provides robustness to scale, rotation, and illumination changes. Compared to Scale-Invariant Feature Transform (SIFT)~\cite{lowe}, AKAZE is significantly faster, making it suitable for real-time applications. While Oriented FAST and Rotated BRIEF (ORB)~\cite{rublee} and Binary Robust Invariant Scalable Keypoints (BRISK)~\cite{Leutenegger} are relatively fast as well, AKAZE generally outperforms them in robustness and distinctiveness. For feature matching, we choose the Brute Force (BF) method to provide high matching values. {The homography matrix $\mathbf{H}_{O \rightarrow C}$, mapping the full workspace pattern to the current camera view, is then calculated using the Random Sample Consensus (RANSAC) algorithm \cite{Fischler1981ransac}, which necessitates a minimum of four matches and effectively excludes outliers for reliable parameter estimation.} Finally, we restore the missing regions in the current camera image by applying $\mathbf{H}_{O \rightarrow C}$ to the full workspace pattern.

\subsection{Relative Pose Error Computation in Simulation}
\label{subsec:pose_error}
After obtaining the restored image, we optimize the probe pose in simulation to minimize the discrepancy between the simulated and restored images. This requires establishing a relationship between image differences and the corresponding pose error. Given the camera's intrinsic matrix $\mathbf{K}$ (assumed identical in both simulation and the real world), our objective is to determine the spatial transformation—comprising the rotation matrix $\mathbf{R}_{\mathcal{G}}^{\mathcal{S}} \in SO(3)$ and translation vector $\mathbf{t}_{\mathcal{G}}^{\mathcal{S}} \in \mathbb{R}^3$—that maps the simulated perspective frame $\{\mathcal{S}\}$ to the real-world perspective frame $\{\mathcal{G}\}$.

Fig. \ref{fig:pipeline} (relative pose error computation part) illustrates this process using the pinhole camera model. Consider a point $\mathbf{X}$ on the textured workspace pattern. In the simulation frame $\{\mathcal{S}\}$, it is represented as $\mathbf{X}^{\mathcal{S}} = [X^{\mathcal{S}}, Y^{\mathcal{S}}, Z^{\mathcal{S}}]^\top$, with image coordinates $\mathbf{p}^{\mathcal{S}} = [u^{\mathcal{S}}, v^{\mathcal{S}}, 1]^\top$. Similarly, in the real-world frame $\{\mathcal{G}\}$, it is expressed as $\mathbf{X}^{\mathcal{G}} = [X^{\mathcal{G}}, Y^{\mathcal{G}}, Z^{\mathcal{G}}]^\top$, with corresponding image coordinates $\mathbf{p}^{\mathcal{G}} = [u^{\mathcal{G}}, v^{\mathcal{G}}, 1]^\top$. The camera projection equations are given by:  
\begin{align}
Z^{\mathcal{G}} \mathbf{p}^{\mathcal{G}} &= \mathbf{K} \mathbf{X}^{\mathcal{G}}, \label{eq_cam_1} \\
Z^{\mathcal{S}} \mathbf{p}^{\mathcal{S}} &= \mathbf{K} \mathbf{X}^{\mathcal{S}}, \label{eq_cam_2}
\end{align}
where the intrinsic matrix $\mathbf{K}$ is defined as:  
\begin{equation}
\mathbf{K} = \begin{bmatrix}
f_x & s & c_x \\
0 & f_y & c_y \\
0 & 0 & 1
\end{bmatrix},
\end{equation}
with $f_x$ and $f_y$ denoting the focal lengths in the x- and y-directions (in pixels), $s$ representing the skew factor, and $(c_x, c_y)$ being the principal point.  

From the spatial transformation in Fig. \ref{fig:pipeline}, the relationship between the two frames is given by:  
\begin{align}
\mathbf{X}^{\mathcal{G}} &= {\mathbf{R}_{\mathcal{G}}^{\mathcal{S}}}^\top (\mathbf{X}^{\mathcal{S}} - \mathbf{t}_{\mathcal{G}}^{\mathcal{S}}), \label{eq_geo_1} \\
d &= -{\mathbf{n}^{\mathcal{S}}}^{\top} \mathbf{X}^{\mathcal{S}}, \label{eq_geo_2}
\end{align}
where $d \in \mathbb{R}^+$ is the shortest distance from the workspace plane to the simulation frame $\{\mathcal{S}\}$, and $\mathbf{n}^{\mathcal{S}} \in \mathbb{R}^3$ is the plane’s normal vector in $\{\mathcal{S}\}$. Combining \eqref{eq_cam_1}, \eqref{eq_cam_2}, \eqref{eq_geo_1}, and \eqref{eq_geo_2}, we derive the homography mapping:  
\begin{equation}
\begin{aligned}
\mathbf{p}^{\mathcal{G}} &=  \xi \mathbf{K} {\mathbf{R}_{\mathcal{G}}^{\mathcal{S}}} ^\top \left(\mathbf{I}_3 + \frac{\mathbf{t}_{\mathcal{G}}^{\mathcal{S}} {\mathbf{n}^{\mathcal{S}}}^\top}{d}\right) \mathbf{K}^{-1} \mathbf{p}^{\mathcal{S}} \\
&= \mathbf{H}_{\mathcal{S} \to \mathcal{G}} \mathbf{p}^{\mathcal{S}},
\end{aligned}
\label{eq:homograph}
\end{equation}  
where $\xi = Z^{\mathcal{S}} / Z^{\mathcal{G}}$ is a scale factor, $\mathbf{I}_3$ is the $3 \times 3$ identity matrix, and $\mathbf{H}_{\mathcal{S} \to \mathcal{G}}$ is the homography matrix mapping frame $\{\mathcal{S}\}$ to $\{\mathcal{G}\}$.

The homography matrix $\mathbf{H}_{\mathcal{S} \to \mathcal{G}}$ can be computed by matching the current simulated image with the restored image using the AKAZE-RANSAC algorithm described in Section \ref{subsec:restoration}. Once obtained, it allows for estimating the 6D pose error between the simulation frame $\{{\mathcal{S}}\}$ and the real-world frame $\{\mathcal{G}\}$. Specifically, $\mathbf{K}$ is obtained via camera calibration, while $\mathbf{n}^{\mathcal{S}}$ and $d$ are known from the simulation. The task then reduces to solving for $\xi$, $\mathbf{R}_{\mathcal{G}}^{\mathcal{S}}$, and $\mathbf{t}_{\mathcal{G}}^{\mathcal{S}}$ given the computed $\mathbf{H}_{\mathcal{S} \to \mathcal{G}}$ from image registration. This is formulated as the following optimization problem:  
\begin{equation}
\min_{\xi, \mathbf{R}, \mathbf{t}} \left\| \mathbf{H}_{\mathcal{S} \to \mathcal{G}} - \xi \mathbf{K} \mathbf{R}^\top \left(\mathbf{I}_3 + \frac{\mathbf{t} {\mathbf{n}^{\mathcal{S}}}^\top}{d}\right) \mathbf{K}^{-1} \right\|_F,
\label{eq:opt}
\end{equation}  
where $\|\cdot\|_F$ denotes the Frobenius norm. Solving this optimization problem provides an accurate estimation of the probe’s pose error between frames $\{\mathcal{S}\}$ and $\{\mathcal{G}\}$, which is then used as a visual feature for the subsequent PBVS control.

{While the homography transformation is mathematically valid only for strictly planar surfaces, minor surface irregularities or slight curvatures are inevitable in practice. Our experiments demonstrate that the proposed image registration and pose estimation pipeline is robust to small deviations from perfect flatness, largely due to feature-based matching and RANSAC outlier rejection. However, substantial deformation or pronounced curvature would violate the planar assumption and introduce localization errors. For clinical or field use, we recommend installing the workspace pattern on a stable, flat substrate to ensure accuracy.}

\subsection{Probe Pose Optimization using Visual Servoing}
\label{subsec:visual_servoing}
We employ a PBVS controller to iteratively minimize the pose error between the simulated and real-world cameras. Thereby, we can obtain the final pose approximation in simulation. To this end, the computed pose error in Section \ref{subsec:pose_error} is used as visual feature, termed as $\mathbf{f} = \left(\mathbf{t}_{\mathcal{G}}^{\mathcal{S}}, \theta \mathbf{u}\right) \in \mathbb{R}^6$, where $\mathbf{t}_{\mathcal{G}}^{\mathcal{S}} \in \mathbb{R}^3$ is the translational component and $\theta\mathbf{u} \in \mathbb{R}^3$ is the axis-angle representation of the rotation matrix $\mathbf{R}_{\mathcal{G}}^{\mathcal{S}}$.  Our control objective is to drive this pose error to zero, i.e., $\mathbf{f}^* = \mathbf{0}$.

The control law is defined as:

\begin{equation}
\mathbf{v} = \mathrm{diag}\{\lambda_t \mathbf{I}_3, \lambda_r \mathbf{I}_3\} \mathbf{e},
\label{eq:vsc}
\end{equation}
where $\mathbf{e} = \mathbf{f} - \mathbf{f}^*$ denotes the visual feature error. The control velocity of the camera, $\mathbf{v} = \left(\mathbf{v}_c, \boldsymbol{\omega}_c\right) \in \mathbb{R}^6$, is expressed in its own frame, where $\mathbf{v}_c \in \mathbb{R}^3$ represents the instantaneous linear velocity of the camera frame’s origin, and $\boldsymbol{\omega}_c \in \mathbb{R}^3$ denotes its instantaneous angular velocity. The translational and rotational gains, $\lambda_t, \lambda_r \in \mathbb{R}^+$, scale the corresponding error components.

Inaccuracies in $\mathbf{f}$ can arise from various sources, such as camera modeling errors and the Sim2Real gap. To balance convergence speed and control stability, we empirically set $\lambda_t$ and $\lambda_r$ within the range of 0 to 1, choosing 0.5 for both in this work. The computed control velocity $\mathbf{v}$ is then applied to the camera in simulation, iteratively guiding the probe toward the desired pose. This process continues until the pose error falls below a predefined threshold of $0.2~\mathrm{mm}$ in translation and $0.2^{\circ}$ in rotation. {These convergence thresholds were empirically selected to provide an optimal trade-off between pose estimation accuracy and computational efficiency: tighter thresholds resulted in only marginal accuracy gains but significantly increased computation time, while looser thresholds caused observable drift in the reconstructed volumes.} Upon convergence, the estimated probe pose in simulation is directly used as an approximation of its real-world counterpart.

\subsection{Sim2Real Calibration}
\label{subsec:compensation}

\begin{figure}[t]
\centering
\includegraphics[width=0.45\textwidth]{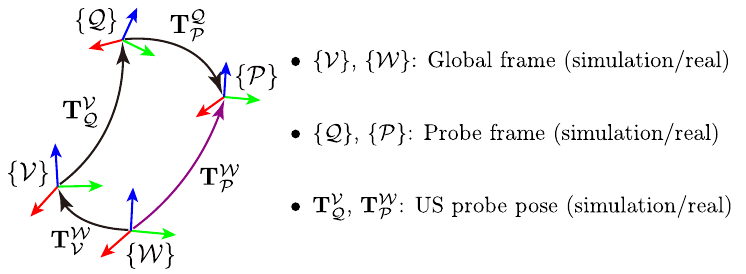}
\caption{{Sim2Real transformation chain. The calibration process estimates the offset transformations $\mathbf{T}_{\mathcal{V}}^{\mathcal{W}}$ and $\mathbf{T}_{\mathcal{P}}^{\mathcal{Q}}$ to align the simulated and real probe poses for accurate 3D US reconstruction.}}
\label{fig:chain}
\end{figure}

Despite the efforts to maintain consistency between simulation and real-world environments, minor discrepancies inevitably arise, as illustrated in Fig. \ref{fig:chain}. Let the US probe pose in simulation be represented by the homogeneous transformation matrix $ \mathbf{T}_{\mathcal{Q}}^{\mathcal{V}} \in SE(3) $ and its actual pose in the real world by $ \mathbf{T}_{\mathcal{P}}^{\mathcal{W}} \in SE(3) $. Here, $ \{\mathcal{V}\} $ and $ \{\mathcal{W}\} $ denote the global frames centered at the workspace pattern in simulation and real-world settings, respectively, while $ \{\mathcal{Q}\} $ and $ \{\mathcal{P}\} $ represent the US probe reference frames in each environment. The relationship between these transformations is given by:

\begin{equation}
\label{eq:sim2real}
\mathbf{T}_{\mathcal{P}}^{\mathcal{W}} = \mathbf{T}_{\mathcal{V}}^{\mathcal{W}} \cdot \mathbf{T}_{\mathcal{Q}}^{\mathcal{V}} \cdot \mathbf{T}_{\mathcal{P}}^{\mathcal{Q}},
\end{equation}
where $\mathbf{T}_{\mathcal{V}}^{\mathcal{W}} \in SE(3)$ represents the pose of the virtual world frame $\{\mathcal{V}\}$ relative to the real-world frame $\{\mathcal{W}\}$, and $\mathbf{T}_{\mathcal{P}}^{\mathcal{Q}}$ captures the misalignment between the real and simulated probe reference frames. Ideally, both transformations should be the identity matrix $\mathbf{I}_4$. However, unavoidable errors, such as workspace misalignment (impacting $\mathbf{T}_{\mathcal{V}}^{\mathcal{W}}$) and hand-eye calibration inaccuracies (affecting $\mathbf{T}_{\mathcal{P}}^{\mathcal{Q}}$), lead to discrepancies between $\mathbf{T}_{\mathcal{P}}^{\mathcal{W}}$ and $\mathbf{T}_{\mathcal{Q}}^{\mathcal{V}}$. Since these offsets remain consistent across different scanning trajectories, we design a calibration process to identify them, as outlined below:

\begin{enumerate}

\item \textbf{Data collection:} Move the probe along predefined trajectories in the real-world setup, recording its actual poses $ \{\mathbf{M}_i\} $, where $\mathbf{M}_i \in SE(3)$ for $i = 1, 2, \dots, l$, along with the corresponding camera images.

\item \textbf{Pose estimation:} Estimate the probe poses $\{\mathbf{N}_i\}$, where $\mathbf{N}_i \in SE(3)$ for $i = 1, 2, \dots, l$, from the captured images using the methods described in Section \ref{sec:method}.

\item \textbf{Transformation computation:} Compute the optimal homogeneous transformation matrices $\mathbf{T}_{\mathcal{V}}^{\mathcal{W}}$ and $\mathbf{T}_{\mathcal{P}}^{\mathcal{Q}}$ to minimize the discrepancy between the estimated and ground-truth poses. This optimization problem is formulated as follows:
\begin{equation} \left(\mathbf{T}_{\mathcal{V}}^{\mathcal{W}}, \mathbf{T}_{\mathcal{P}}^{\mathcal{Q}}\right) = \arg\min_{\mathbf{U}, \mathbf{V} \in SE(3)} \sum_{i=1}^{l} \mathcal{L}_i\left(\mathbf{U}, \mathbf{V} \right), \end{equation}
where the loss function $\mathcal{L}_i$ is defined as: \begin{equation} \mathcal{L}_i = \left\|{\mathbf{t}_m}_i - {\mathbf{t}_n}_i\right\| + \zeta \cos^{-1} \left( \frac{ \operatorname{Tr}({{\mathbf{R}^{\top}_m}_i} {\mathbf{R}_n}_i) - 1}{2} \right). \end{equation} 
Here, $\operatorname{Tr}(\cdot)$ denotes the matrix trace, and the pose components ${\mathbf{R}_m}_i, {\mathbf{R}_n}_i \in SO(3)$, ${\mathbf{t}_m}_i, {\mathbf{t}_n}_i \in \mathbb{R}^3$ are obtained from: 
\begin{equation}
   \mathbf{M}_i = \begin{bmatrix}
       {\mathbf{R}_m}_i & {\mathbf{t}_m}_i \\
       \mathbf{0}^{\top} & 1
   \end{bmatrix}, \quad
   \mathbf{U} \mathbf{N}_i \mathbf{V} = \begin{bmatrix}
       {\mathbf{R}_n}_i & {\mathbf{t}_n}_i \\
       \mathbf{0}^{\top} & 1
   \end{bmatrix}.
\end{equation}
{The scaling factor $\zeta \in \mathbb{R}$ was empirically set to 0.1 to balance translational and rotational errors, so that neither component dominates the optimization and stable convergence is achieved in practice.}

\end{enumerate}

Once computed, the transformations $\mathbf{T}_{\mathcal{V}}^{\mathcal{W}}$ and $\mathbf{T}_{\mathcal{P}}^{\mathcal{Q}}$ are applied to correct the estimated probe pose $\mathbf{T}_{\mathcal{Q}}^{\mathcal{V}}$ according to (\ref{eq:sim2real}), effectively bridging the Sim2Real gap. Importantly, this calibration procedure needs to be performed only once provided that the system configuration—namely, workspace pattern and camera configuration—remains unchanged, thereby imposing minimal overhead for subsequent US scans.

\subsection{Dual-Camera Setting}
\label{subsec:multi_camera}
{When the camera's FoV is fully occluded, no matching keypoints can be detected in the current image, making it impossible to estimate the probe pose. To overcome this limitation, we utilize two eye-in-hand cameras to ensure reliable pose estimation.} When both cameras are operational, the probe pose is computed as a weighted average of the pose estimates derived from each camera's images. Given that the final visual feature error from a single camera is defined as \(\mathbf{e}^* = \left({\mathbf{t}_{\mathcal{G}}^{\mathcal{S}}}^*, {\theta \mathbf{u}}^*\right)\), the weight assigned to its pose estimation is given by:
\begin{equation}
w = 1/\left(1 + \mu_t \|{\mathbf{t}_{\mathcal{G}}^{\mathcal{S}}}^*\| + \mu_r \|{\theta \mathbf{u}}^*\|\right),
\label{eq:weight}
\end{equation}
{where $\mu_t, \mu_r \in \mathbb{R}$ were empirically set to 1000 and 50, respectively, to ensure comparable contributions from translation and rotation errors in the weighting function.} {This inverse weighting scheme ensures that a smaller residual error leads to a larger weight, adaptively prioritizing the camera that provides a more reliable pose estimate in each frame.}

Let the weights for the left and right cameras be denoted as \(w_l\) and \(w_r\), respectively. Their estimated probe poses are represented as \((\mathbf{t}_l, \mathbf{q}_l)\) and \((\mathbf{t}_r, \mathbf{q}_r)\), where \(\mathbf{t}_l, \mathbf{t}_r \in \mathbb{R}^3\) are translation vectors, and \(\mathbf{q}_l, \mathbf{q}_r\) are quaternion vectors. The weighted average translation vector is computed as:
\begin{equation}
\label{eq:t_mean}
\mathbf{t}_{avg} = \frac{w_l \mathbf{t}_l + w_r \mathbf{t}_r}{w_l + w_r}.
\end{equation}
The weighted average quaternion is obtained using spherical linear interpolation (Slerp) \cite{shoemake1985animating}:
\begin{equation}
\label{eq:q_mean}
\mathbf{q}_{avg} = \text{Slerp}\left(\mathbf{q}_l, \mathbf{q}_r; \frac{w_r}{w_l + w_r}\right),
\end{equation}
where \(\text{Slerp}(\cdot)\) ensures smooth interpolation between quaternions. Thus, the combined weighted average probe pose is represented by \((\mathbf{t}_{avg}, \mathbf{q}_{avg})\). In cases where only one camera is operational, the probe pose directly corresponds to the estimate from the functioning camera. If neither camera is operational, no pose estimate is generated.

\subsection{Pseudocode for Probe Pose Estimation}
\label{subsec:pseudocode}
To illustrate the implementation details of the proposed probe pose estimation method, we provide the pseudocode in \textbf{Algorithm} \ref{alg:process}, specifically demonstrating the pose estimation process using the left camera. Before executing the pose estimation algorithm, it is crucial to establish a simulation environment in CoppeliaSim. This setup involves aligning the camera model, object geometry, and image pattern with their real-world counterparts to ensure accuracy and realism. Additionally, to ensure that the restored real-world observation is contained within the initial simulation image, the probe pose in the simulation is randomly initialized to make sure the left camera's FoV completely captures the workspace pattern. By following the procedure described in \textbf{Algorithm} \ref{alg:process}, an optimal estimate of the US probe pose, denoted by $\mathbf{T}_l$, can be obtained, along with its confidence weight $w_l$. Similarly, observations from the right camera can yield another precise estimation of the US probe pose, represented as $\mathbf{T}_r$ with confidence weight $w_r$. Finally, the overall probe pose leveraging dual-camera data is computed using \eqref{eq:t_mean} and \eqref{eq:q_mean}.

\begin{algorithm}[t]
\caption{Probe Pose Estimation via Left Camera}
\label{alg:process}
\small
\KwIn{Real-world left camera image $\mathbf{I}_{l}$, workspace pattern image $\mathbf{I}_{ws}$.}
\KwOut{Estimated US probe pose $\mathbf{T}_{l} \in SE(3)$, confidence weight $w_{l} \in \left(0, 1\right)$.}
\mycomment{\tcc{\textbf{Stage 1: Sim2Real Calibration}}}
Acquire calibration transformations $\left({\mathbf{T}_{\mathcal{V}}^{\mathcal{W}}}_{l}, {\mathbf{T}_{\mathcal{P}}^{\mathcal{Q}}}_{l}\right)$ for the left camera as detailed in Section~\ref{subsec:compensation}\;

Initialize $\mathbf{T}_{l} \leftarrow \text{None}$, $\mathbf{\hat{T}}_{l} \leftarrow \mathbf{0}_{4 \times 4}$, $w_{l} \leftarrow 0$\;

\mycomment{\tcc{\textbf{Stage 2: Recover Standardized Camera Observation}}}
Compute homography $\mathbf{H}_{O \rightarrow C}$ mapping $\mathbf{I}_{ws}$ to $\mathbf{I}_{l}$\;

\If{$\mathbf{H}_{O \rightarrow C}$ computation fails}{
   \Return{$\mathbf{T}_{l}$, $w_{l}$}\;
}

Warp $\mathbf{I}_{ws}$ using $\mathbf{H}_{O \rightarrow C}$ to get restored image $\mathbf{I}^{\prime}_{l}$\;

Set $k \leftarrow 0$, $\text{max\_iter} \leftarrow 20$\;

\mycomment{\tcc{\textbf{Stage 3: Iterative Pose Optimization}}}
\Repeat{$\left(\left\|\mathbf{t}_{\mathcal{G}}^{\mathcal{S}}\right\| < 0.2~\mathrm{mm} \land \left\|\theta \mathbf{u}\right\| < 0.2^{\circ}\right) \lor k \geq \text{max\_iter}$}{
    Acquire simulated left camera image $\mathbf{L}_{l}$\;

    Compute homography $\mathbf{H}_{\mathcal{S} \rightarrow \mathcal{G}}$ mapping $\mathbf{L}_{l}$ to $\mathbf{I}^{\prime}_{l}$\;

    \If{$\mathbf{H}_{\mathcal{S} \rightarrow \mathcal{G}}$ computation fails}{
        Set $\mathbf{\hat{T}}_{l} \leftarrow \text{None}$ and \textbf{break}\;
    }

    Estimate pose error ($\mathbf{R}_{\mathcal{G}}^{\mathcal{S}}$, $\mathbf{t}_{\mathcal{G}}^{\mathcal{S}}$) using \eqref{eq:opt}, and compute feature error $\mathbf{e} = \left(\mathbf{t}_{\mathcal{G}}^{\mathcal{S}}, \theta \mathbf{u}\right)$\;
    
    Compute desired camera velocity $\mathbf{v}$ using \eqref{eq:vsc}\;

    Update simulated left camera pose $\mathbf{\hat{T}}_{l}$ by applying velocity $\mathbf{v}$ over time step $\Delta t$\;

    Increment counter $k \leftarrow k + 1$\;
}

\mycomment{\tcc{\textbf{Stage 4: Sim2Real Compensation and Output}}}
\If{$\mathbf{\hat{T}}_{l}$ is not None}{
    Retrieve probe pose $\mathbf{\hat{T}}_{l} \in SE(3)$ from simulation\;
    
    Compute final probe pose $\mathbf{T}_{l} \leftarrow {\mathbf{T}_{\mathcal{V}}^{\mathcal{W}}}_{l} \cdot \mathbf{\hat{T}}_{l} \cdot {\mathbf{T}_{\mathcal{P}}^{\mathcal{Q}}}_{l}$\;

    Calculate confidence weight $w_{l}$ using \eqref{eq:weight}\;
}

\Return{$\mathbf{T}_{l}$, $w_{l}$}\;\label{return}

\end{algorithm}

\section{Experiments and Results}
\label{sec:exp}
This section validates the effectiveness of the proposed US probe pose estimation algorithm. First, we present an intuitive demonstration of the image restoration method in real-world scenarios, addressing practical challenges such as occlusions, reflections, and low illumination. Next, we quantitatively evaluate the pose estimation performance by maneuvering the probe along U-shaped and spiral trajectories. Finally, we showcase the method's effectiveness in 3D reconstruction through freehand scanning of various subjects, including a vascular phantom, a 3D-printed conical model, and a human arm.

\subsection{Image Restoration Performance}
\label{subsec:restore_perf}
\par
\begin{figure}[htp]
    \centering 
    \begin{minipage}{0.2\textwidth}
        \centering    
        \begin{tikzpicture}
            \node[anchor=south west, inner sep=0] (image) at (0,0) 
                {\includegraphics[width=\textwidth]{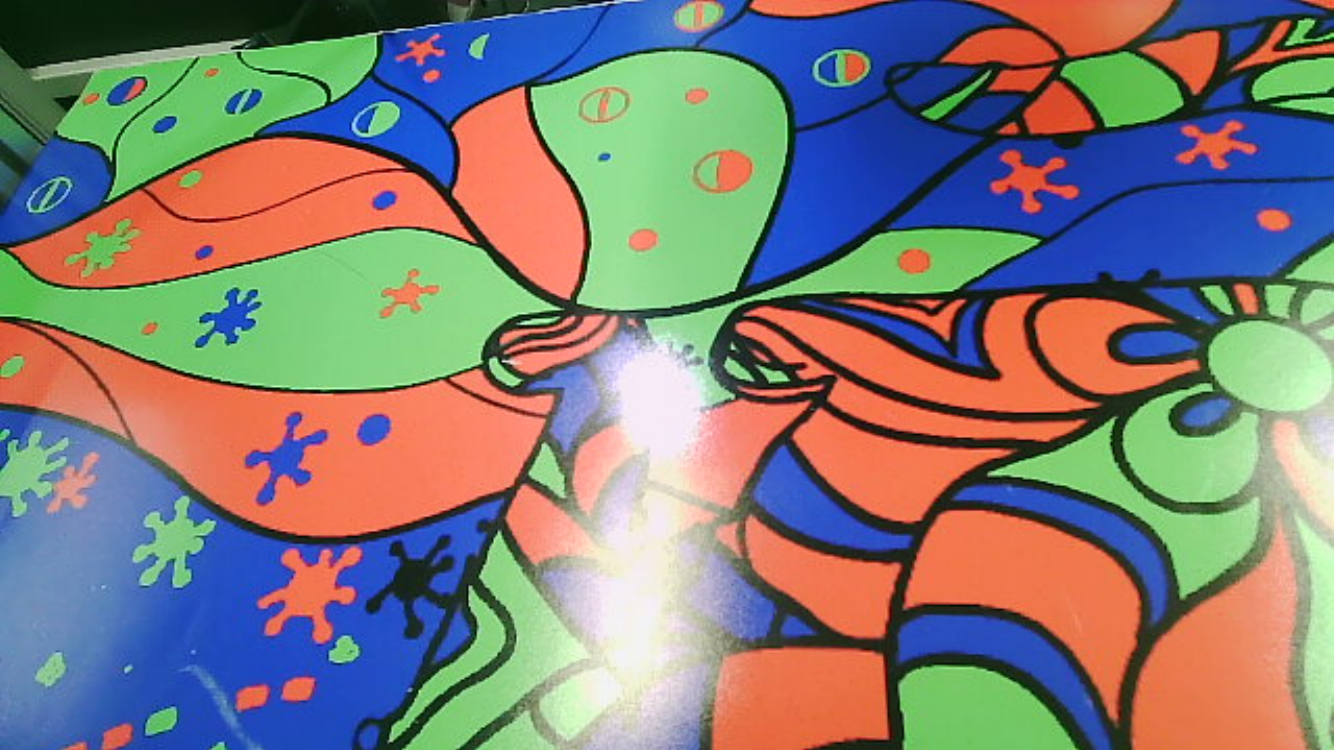}};
            \node[fill=white, draw=black, inner sep=1pt] at (0.3, 1.8) {(a)};
        \end{tikzpicture}
    \end{minipage}
    \begin{minipage}{0.2\textwidth}
        \centering
        \begin{tikzpicture}
            \node[anchor=south west, inner sep=0] (image) at (0,0) 
                {\includegraphics[width=\textwidth]{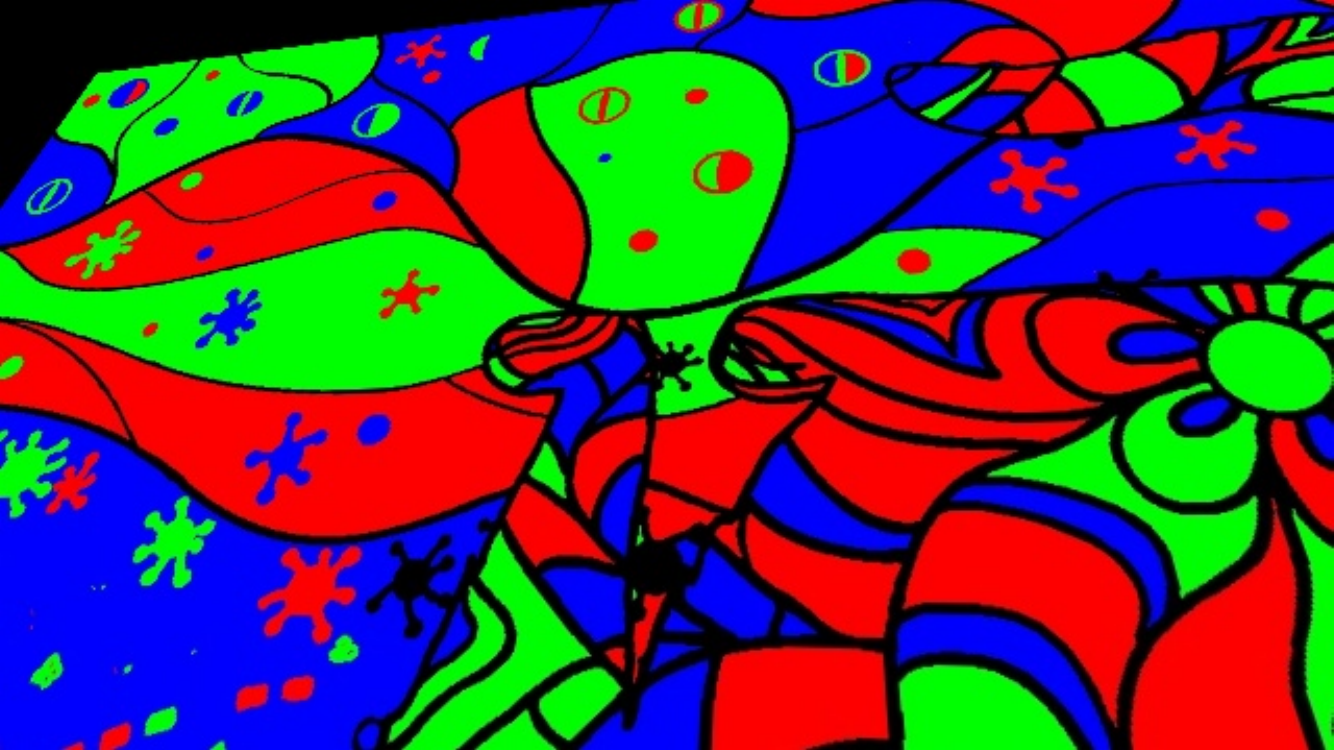}};
            \node[fill=white, draw=black, inner sep=1pt] at (0.3, 1.8) {(d)};
        \end{tikzpicture}
    \end{minipage}

    \begin{minipage}{0.2\textwidth}
        \centering
        \begin{tikzpicture}
            \node[anchor=south west, inner sep=0] (image) at (0,0) 
                {\includegraphics[width=\textwidth]{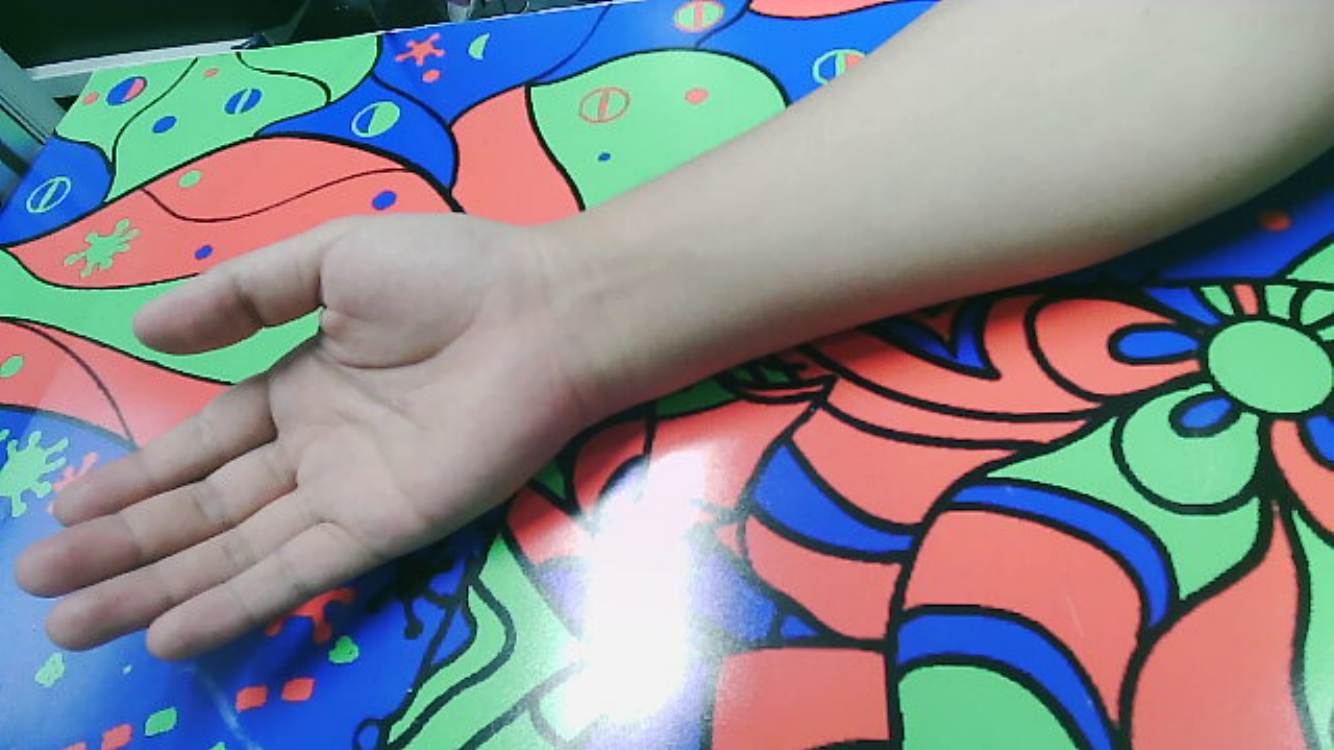}};
            \node[fill=white, draw=black, inner sep=1pt] at (0.3, 1.8) {(b)};
        \end{tikzpicture}
    \end{minipage}
    \begin{minipage}{0.2\textwidth}
        \centering
        \begin{tikzpicture}
            \node[anchor=south west, inner sep=0] (image) at (0,0) 
                {\includegraphics[width=\textwidth]{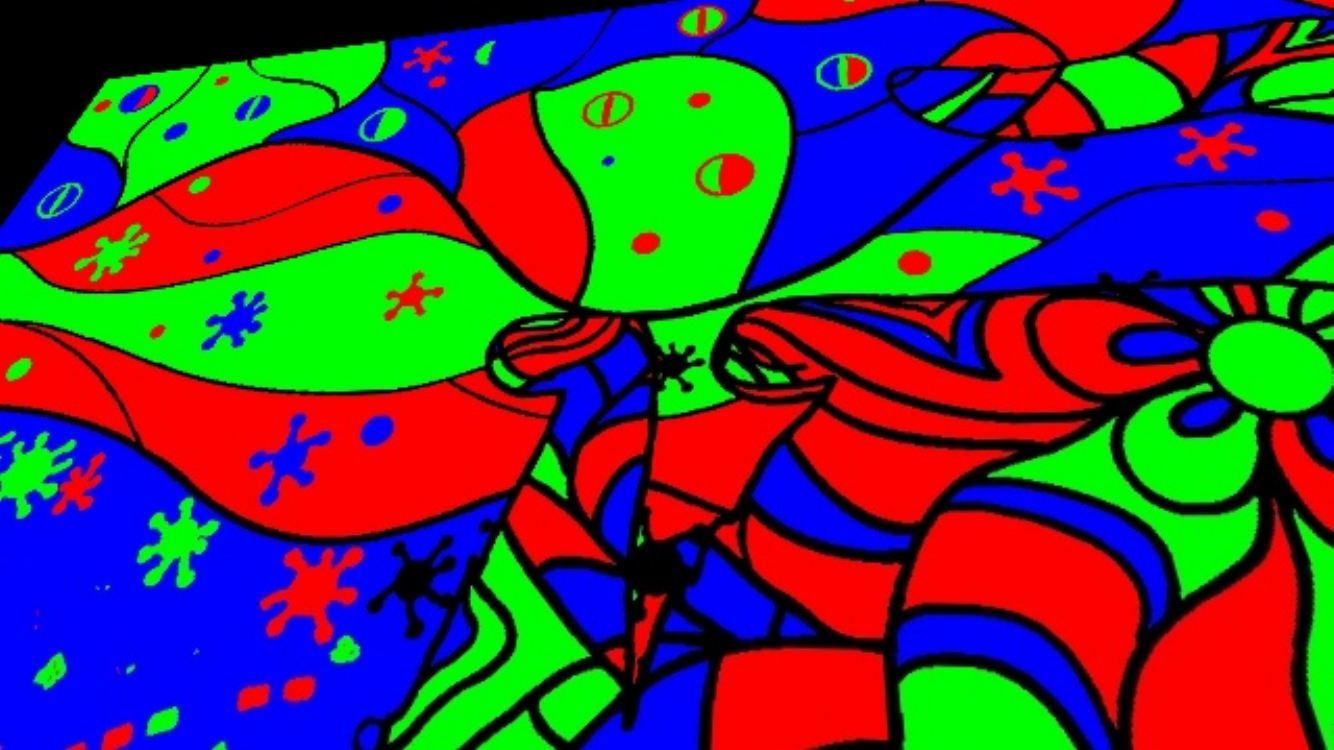}};
            \node[fill=white, draw=black, inner sep=1pt] at (0.3, 1.8) {(e)};
        \end{tikzpicture}
    \end{minipage}

    \begin{minipage}{0.2\textwidth}
        \centering
        \begin{tikzpicture}
            \node[anchor=south west, inner sep=0] (image) at (0,0) 
                {\includegraphics[width=\textwidth]{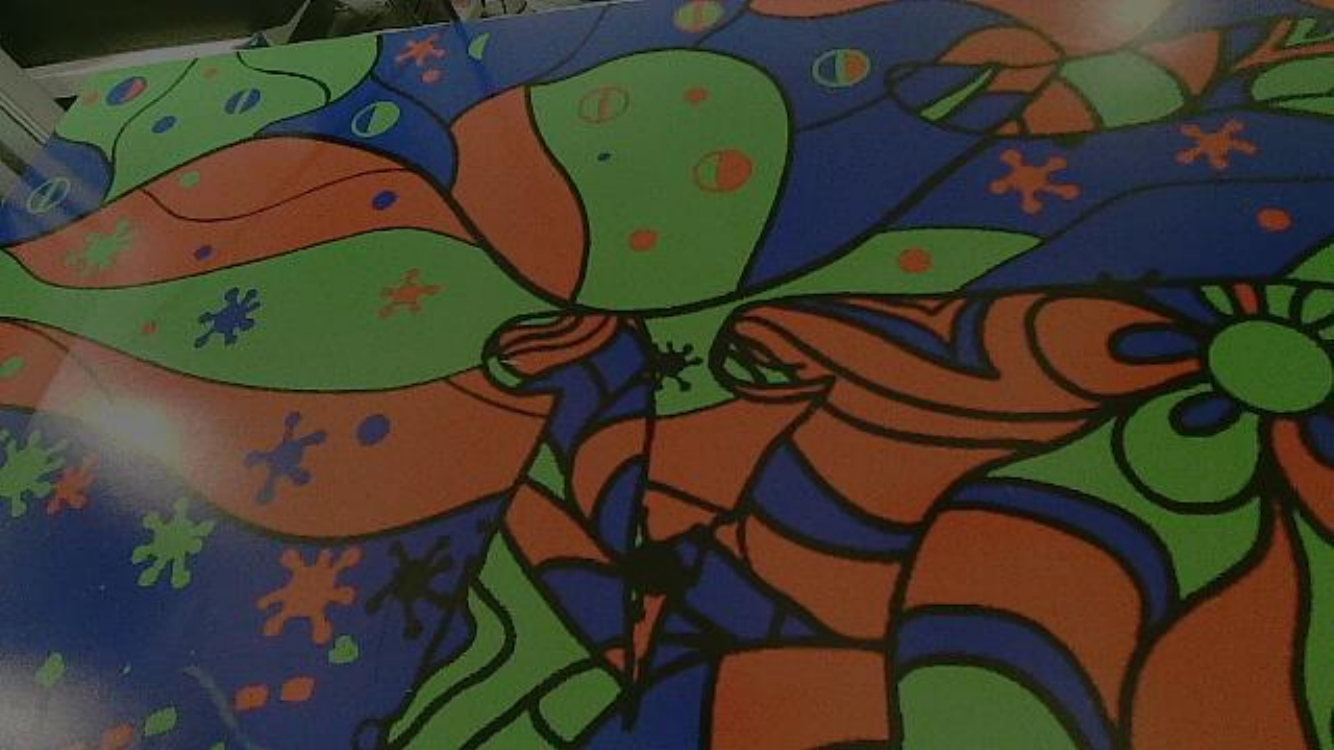}};
            \node[fill=white, draw=black, inner sep=1pt] at (0.3, 1.8) {(c)};
        \end{tikzpicture}
    \end{minipage}
    \begin{minipage}{0.2\textwidth}
        \centering
        \begin{tikzpicture}
            \node[anchor=south west, inner sep=0] (image) at (0,0) 
                {\includegraphics[width=\textwidth]{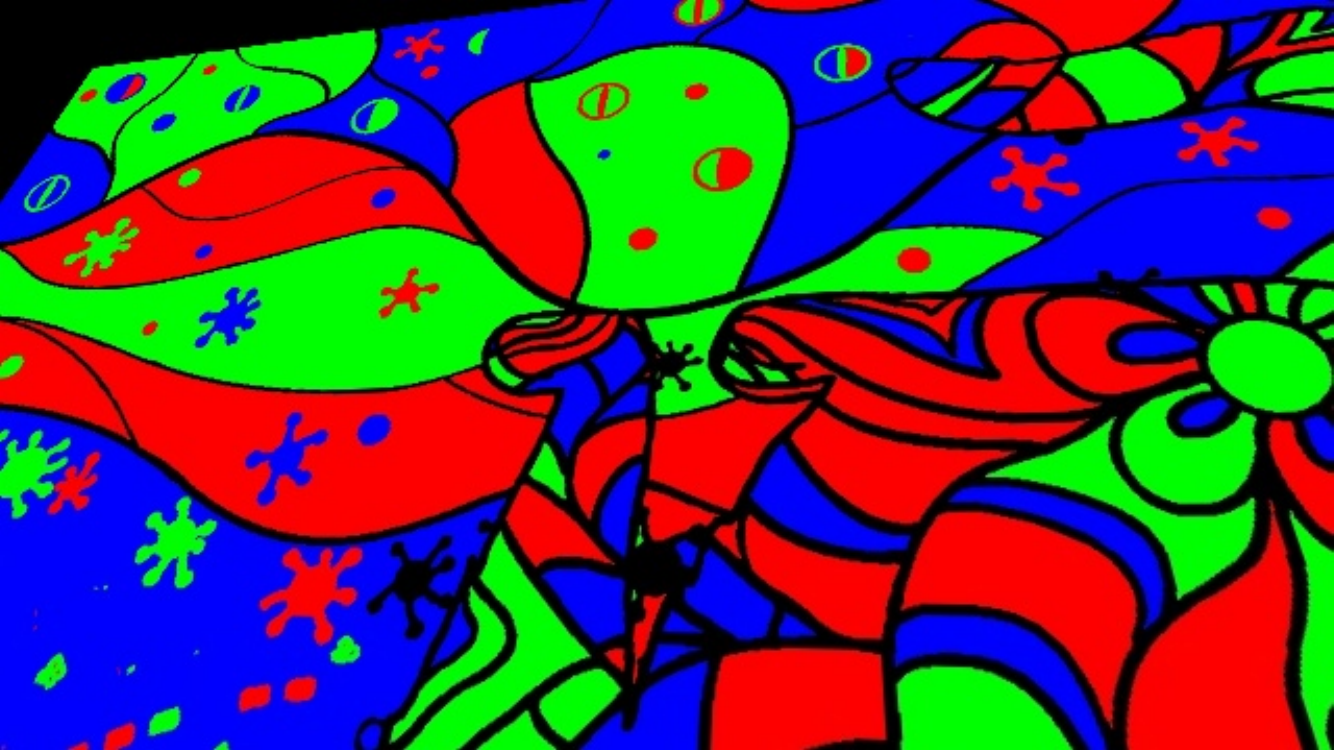}};
            \node[fill=white, draw=black, inner sep=1pt] at (0.3, 1.8) {(f)};
        \end{tikzpicture}
    \end{minipage}

    \caption{Image restoration results: (a) image affected by lighting reflection, (b) image with arm occlusion, and (c) image captured in dim lighting. (d), (e), and (f) show the restored versions of (a), (b), and (c), respectively.}
    \label{fig:exp_rst}
\end{figure}
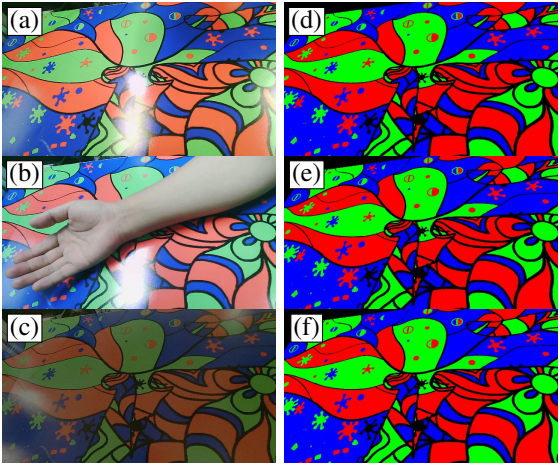

In this study, we propose an image restoration algorithm to mitigate the negative effects of variations in camera observations by leveraging prior knowledge of the workspace pattern. To evaluate its effectiveness, we examine three challenging scenarios that commonly arise in freehand US probe pose estimation: (a) strong lighting reflections in the camera’s FoV due to excessive external illumination, (b) occlusions caused by the presence of an examined target, such as a human arm, and (c) dim lighting conditions due to insufficient illumination. An intuitive visualization is provided in Fig.~\ref{fig:exp_rst}. The proposed restoration algorithm effectively recovers image details obscured by these challenging conditions. The restored images exhibit clear and consistent RGB patterns, demonstrating robustness in visual appearance. By reducing discrepancies between simulated and real-world images, the restoration process can further enhance the stability and precision of the proposed pose estimation algorithm.

\subsection{Probe Pose Estimation Performance}
\label{subsec:traj_eval}
\begin{figure}[t]
    \centering
    \begin{minipage}{0.24\textwidth}
        \centering    
        \begin{tikzpicture}
            \node[anchor=south west, inner sep=0] (image) at (0,0) 
                {\includegraphics[width=\textwidth]{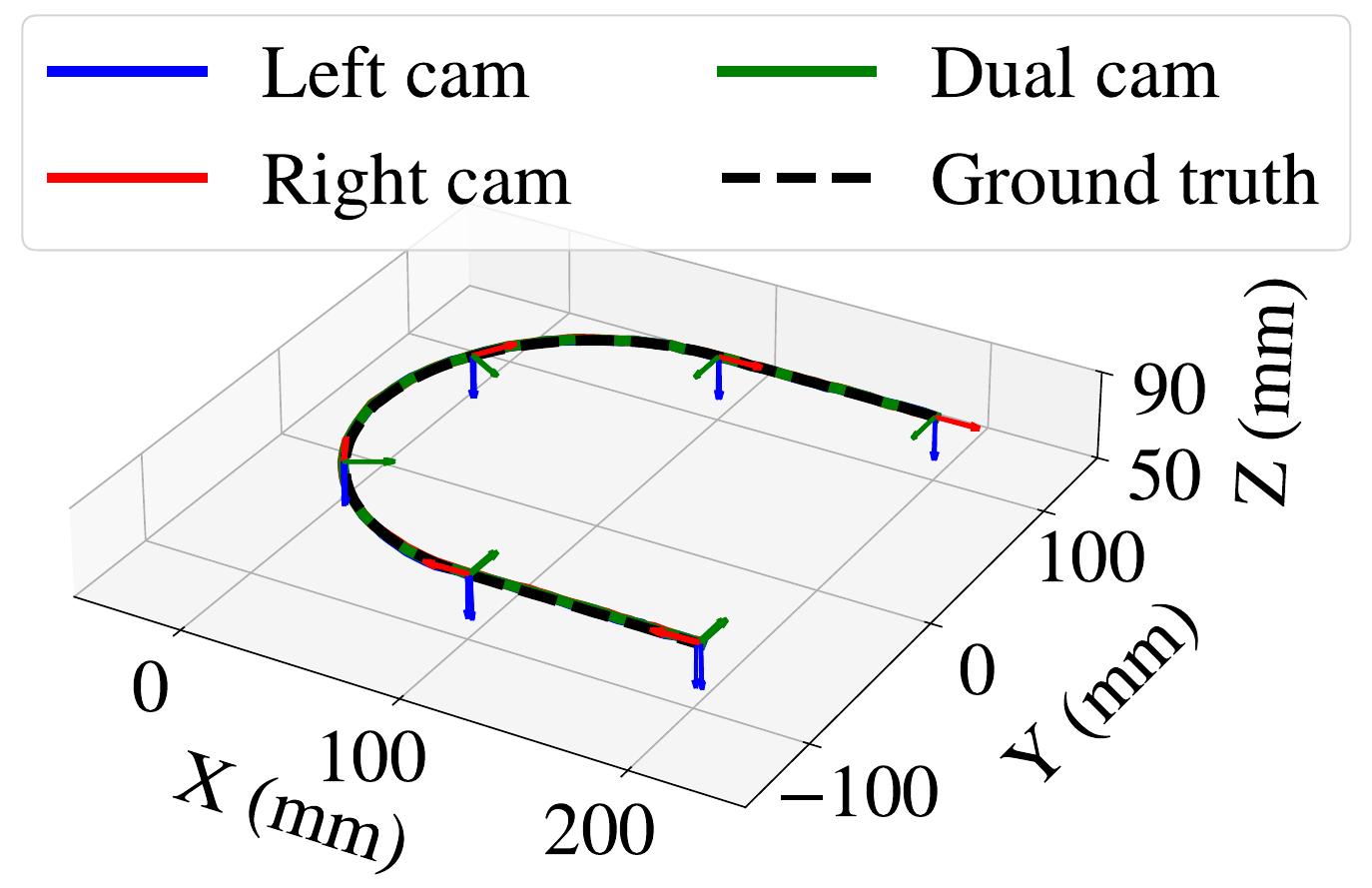}};
            \node[inner sep=1pt] at (0.4, 1.8) {(a)};
        \end{tikzpicture}
    \end{minipage}
    \begin{minipage}{0.24\textwidth}
        \centering
        \begin{tikzpicture}
            \node[anchor=south west, inner sep=0] (image) at (0,0) 
                {\includegraphics[width=\textwidth]{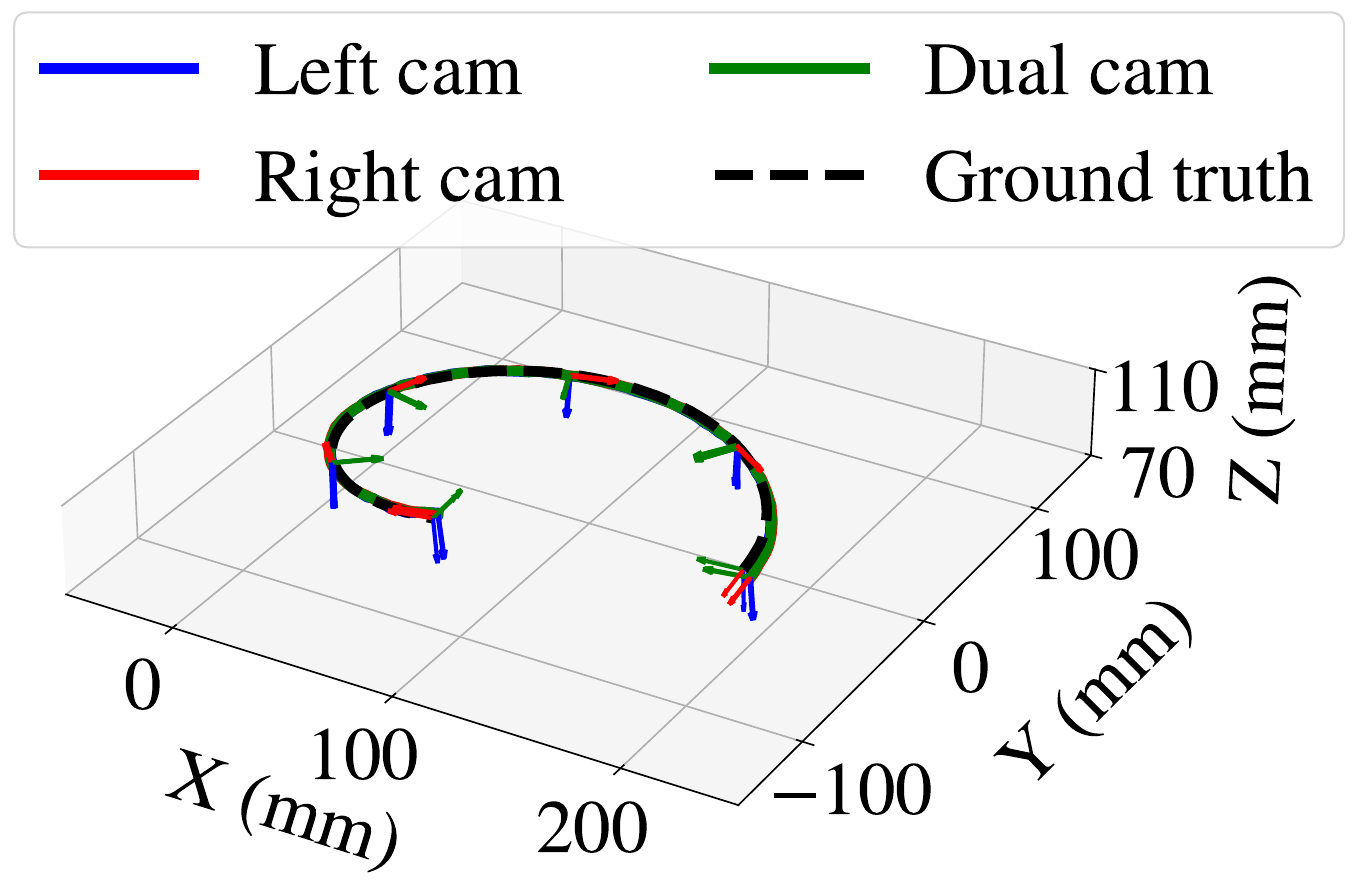}};
            \node[inner sep=1pt] at (0.4, 1.8) {(b)};
        \end{tikzpicture}
    \end{minipage}
    \caption{Trajectory Estimation Results: (a) U-shaped, (b) Spiral. US probe trajectories are estimated using the left camera (blue), right camera (red), and both cameras (green).}
    \label{fig:exp_traj}
\end{figure}

To quantitatively evaluate the proposed method in real-world scenarios, we utilize a robotic arm to maneuver the probe along a predefined U-shaped trajectory and a spiral trajectory in 3D space (see Fig.~\ref{fig:exp_traj}). 
The U-shape trajectory is a planar path parallel to the workspace, which combines two $100~\mathrm{mm}$ straight segments and a semicircular arc with a $100~\mathrm{mm}$ radius at a fixed height of $80~\mathrm{mm}$. To further evaluate the estimation performance in full 3D space, a spiral trajectory is designed, which has a $3/4$ circular arc with a radius decreasing from $100~\mathrm{mm}$ to $70~\mathrm{mm}$ and a height range of $70~\mathrm{mm}$ to $100~\mathrm{mm}$. These two trajectories span a relatively large workspace, which can ensure the proposed method's applicability for scanning the human body. It is worth noting that the robot is mainly used to follow these scanning trajectories accurately. {Simultaneously, image streams from two low-cost cameras are recorded to compute the probe pose, while the robotic tracking data serves as the ground truth. Both camera images and ground-truth probe poses are recorded and software-synchronized using ROS.}

\begin{figure}[htp]
    \centering
    \begin{minipage}{0.45\textwidth}
        \centering    
        \begin{tikzpicture}
            \node[anchor=south west, inner sep=0] (image) at (0, 0) 
                {\includegraphics[width=\textwidth]{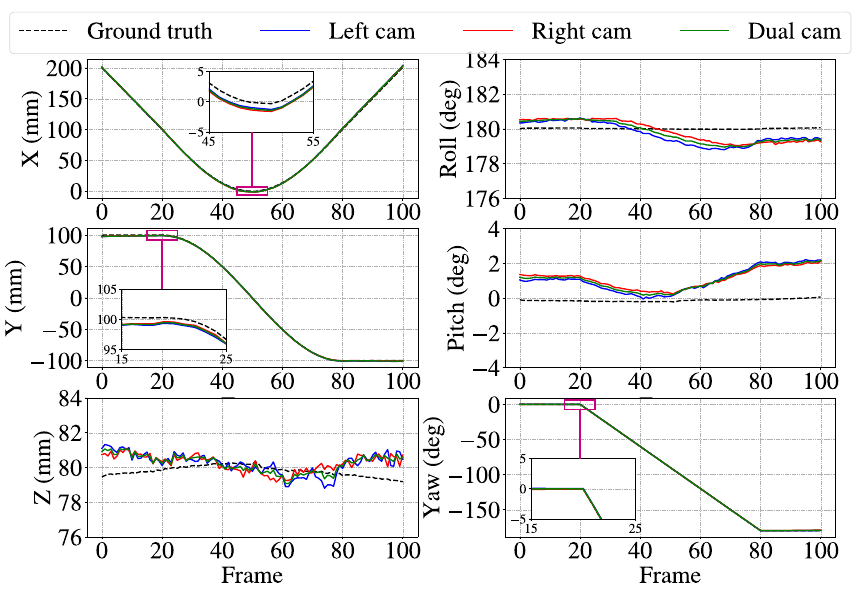}};
            \node[black, anchor=north] at (0.0, 5.2) {(a)};
        \end{tikzpicture}
    \end{minipage}
    \begin{minipage}{0.44\textwidth}
        \centering
        \begin{tikzpicture}
            \node[anchor=south west, inner sep=0] (image) at (0.3, 0) 
                {\includegraphics[width=\textwidth]{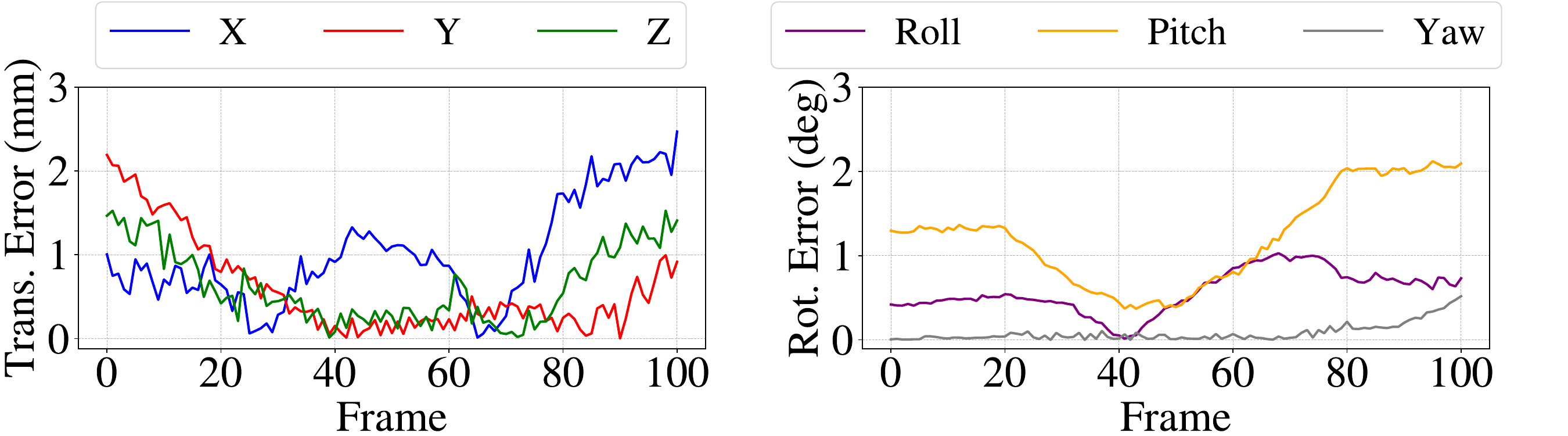}};
            \node[black, anchor=north] at (0.0, 2.0) {(b)};
        \end{tikzpicture}
    \end{minipage}
    \caption{{Performance analysis for the U-shaped trajectory: (a) Translational and rotational predictions for left, right, and dual-camera configurations; (b) 6-DoF pose estimation errors for the dual-camera configuration.}}
    \label{fig:u_traj}
\end{figure}

\begin{figure}[htp]
    \centering
    \begin{minipage}{0.45\textwidth}
        \centering
        \begin{tikzpicture}
            \node[anchor=south west, inner sep=0] (image) at (0, 0) 
                {\includegraphics[width=\textwidth]{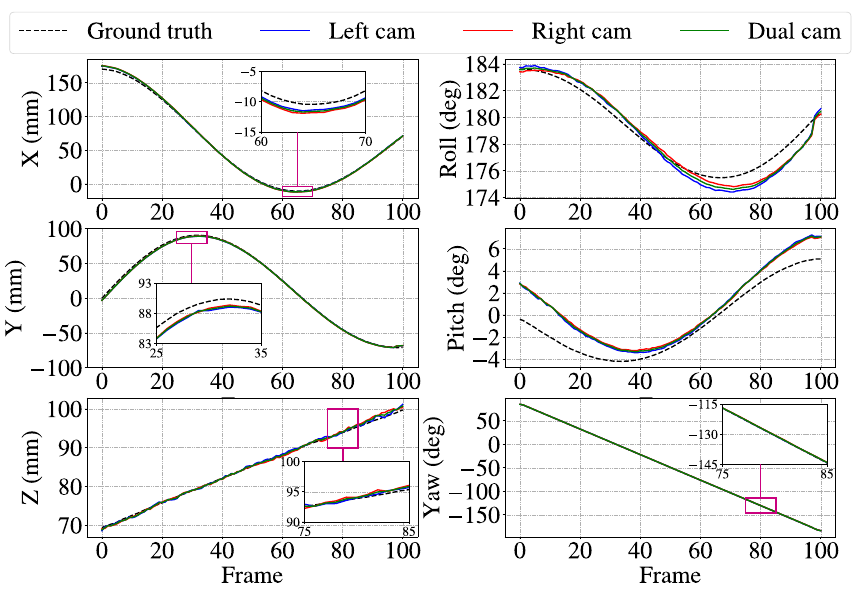}};
            \node[black, anchor=north] at (0.0, 5.2) {(a)};
        \end{tikzpicture}
    \end{minipage}
    \begin{minipage}{0.44\textwidth}
        \centering
        \begin{tikzpicture}
            \node[anchor=south west, inner sep=0] (image) at (0.3, 0) 
                {\includegraphics[width=\textwidth]{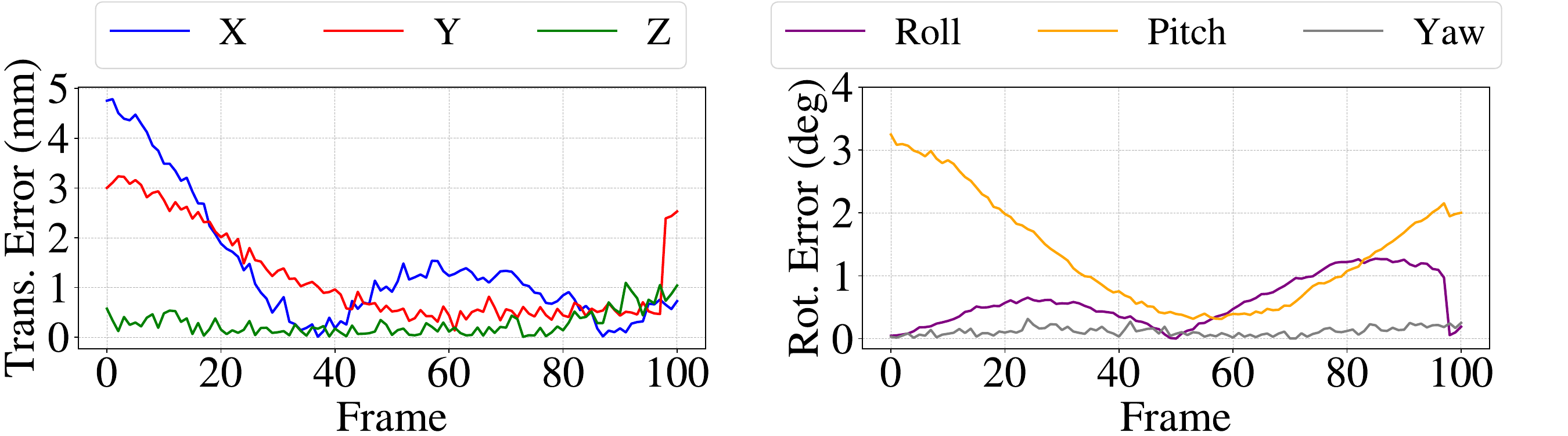}};
            \node[black, anchor=north] at (0.0, 2.0) {(b)};
        \end{tikzpicture}
    \end{minipage}
    \caption{{Performance analysis for the spiral trajectory: (a) Translational and rotational predictions for left, right, and dual-camera configurations; (b) 6-DoF pose estimation errors for the dual-camera configuration.}}
    \label{fig:spi_traj}
\end{figure}

The intuitive visualization of probe pose estimation is shown in Fig.~\ref{fig:exp_traj}. The figure demonstrates that the proposed camera-based pose estimation method can accurately predict the probe pose across different settings (i.e., using only the left camera, only the right camera, or both cameras). To further evaluate axis-specific performance, we decompose the estimation errors along the scanning trajectory into individual translational and rotational components (see Figs. \ref{fig:u_traj} and \ref{fig:spi_traj} for the U-shaped and spiral trajectories, respectively). Statistical results for these trajectories are summarized in Table~\ref{tab:traj_err}. For the U-shaped trajectory, the average translational and rotational errors are $1.460~\mathrm{mm}$ and $1.368^{\circ}$, respectively, while for the spiral trajectory, they increase slightly to $1.975~\mathrm{mm}$ and $1.565^{\circ}$.
{In addition, the maximum drift refers to the largest instantaneous deviation between the estimated pose and the ground truth along the scanning trajectory.} Notably, the maximum drift observed in both U-shaped and spiral trajectories remains within a reasonable boundary, indicating the stability of our method over long scanning paths.
We need to emphasize that the dual-camera configuration is primarily intended to enhance method robustness by mitigating severe occlusions occurring from specific directions. The fusion procedure described in Section \ref{subsec:multi_camera} integrates the estimation results obtained independently from each camera. Consequently, as indicated in Table \ref{tab:traj_err}, the dual-camera setup does not always yield the best performance among the three configurations. Nevertheless, the estimation results remain highly consistent across different camera arrangements (e.g., left versus right camera), highlighting the robustness and scalability of the proposed method for diverse applications.

\begin{table*}[t]
\centering
\caption{Comparison of Translational and Rotational Errors in US Probe Pose Predictions}
\label{tab:traj_err}
\resizebox{0.65\textwidth}{!}{
\begin{tabular}{lllcccc}
\toprule
 \multirow{2}{*}{\textbf{Trajectory}} & \multirow{2}{*}{\textbf{Method}} & \multirow{2}{*}{\textbf{Camera}} & \multicolumn{2}{c}{\textbf{Avg. Error (mm/deg)}} & \multicolumn{2}{c}{\textbf{Max. Drift (mm/deg)}} \\ 
\cmidrule(lr){4-5} \cmidrule(lr){6-7}
       &                     &                         & \textbf{Trans. $\downarrow$} & \textbf{Rot. $\downarrow$}  & \textbf{Trans. $\downarrow$} & \textbf{Rot. $\downarrow$} \\ 
\midrule

\multirow{12}{*}{\begin{tabular}[c]{@{}l@{}} U-shaped \end{tabular}}
& \multirow{3}{*}{Dense AprilTag} & Left only & 2.097 $\pm$ 0.914 & 1.458 $\pm$ 0.908   & 4.324 & 2.787                                              \\
                       &  & Right only & 2.213 $\pm$ 1.025                                                  & 1.422 $\pm$ 0.480    &   4.718 & 2.148                                          \\
                       &  & Dual       & 2.120 $\pm$ 0.974                                                  & 1.393 $\pm$ 0.666    & 4.421 & 2.428                                             \\ \cmidrule(lr){2-7}
& \multirow{3}{*}{{ORB-SLAM3}} & {Left only} & {5.886 $\pm$ 1.295} & {4.829 $\pm$ 0.998}   & {12.697} & {6.844}                                           \\
                       &  & {Right only} & {6.314 $\pm$ 1.515}                                                  & {3.826 $\pm$ 0.877}    &   {14.721} & {6.544}                                          \\
                       &  & {Dual}       & {6.217 $\pm$ 1.402}                                                  & {3.923 $\pm$ 0.914}    & {13.683} & {6.502}                                             \\ \cmidrule(lr){2-7}
& \multirow{3}{*}{\begin{tabular}[c]{@{}l@{}} Our Method w/o\\ Visual Servoing \end{tabular}} & Left only  & 1.747 $\pm$ 0.871  & 1.521 $\pm$ 0.507 & 6.683 & 2.361 \\
                        &                         & Right only & 1.575 $\pm$ 0.750  & 1.538 $\pm$ 0.439 & 5.109 & 2.346  \\
                        &                         & Dual       & 1.614 $\pm$ 0.812  & 1.512 $\pm$ 0.475 & 5.865 & 2.322 \\ \cmidrule(lr){2-7}
& \multirow{3}{*}{Our Method} & Left only  & 1.530 $\pm$ 0.808  & 1.363 $\pm$ 0.611 & 3.391 & 2.288 \\
                        &                         & Right only & 1.453 $\pm$ 0.605  & 1.404 $\pm$ 0.537 & 2.654 & 2.357  \\
                        &                         & Dual       & \textbf{1.460 $\pm$ 0.696}  & \textbf{1.368 $\pm$ 0.574} & \textbf{2.988} & \textbf{2.275} \\ \cmidrule(lr){1-7}
\multirow{12}{*}{\begin{tabular}[c]{@{}l@{}} Spiral \end{tabular}}  & \multirow{3}{*}{Dense AprilTag}  & Left only  & 3.120 $\pm$ 1.695                                                  & 1.600 $\pm$ 0.814  & 6.929 & 2.845                                               \\
                       &  & Right only & 2.913 $\pm$ 1.363                                                  & 1.561 $\pm$ 0.730   & 5.696 & 3.076                                              \\
                       &  & Dual       & 2.938 $\pm$ 1.561                                                  & \textbf{1.514 $\pm$ 0.752}   & 6.267 & \textbf{2.874}                                              \\ \cmidrule(lr){2-7}
& \multirow{3}{*}{{ORB-SLAM3}} & {Left only} & {7.345 $\pm$ 1.436} & {5.160 $\pm$ 0.975}   & {13.736} & {7.177}\\
                       &  & {Right only} & {8.209 $\pm$ 2.018}                                                  & {4.309 $\pm$ 0.980}    &   {15.253} & {6.907}                                          \\
                       &  & {Dual}       & {8.041 $\pm$ 1.427}                                                  & {4.419 $\pm$ 0.814}    & {13.766} & {7.023}                                             \\ \cmidrule(lr){2-7}
& \multirow{3}{*}{\begin{tabular}[c]{@{}l@{}} Our Method w/o\\ Visual Servoing \end{tabular}} & Left only  & 2.156 $\pm$ 1.904                                                  & 1.751 $\pm$ 0.701   & 7.474 & 3.245                                              \\
                       &  & Right only & 2.359 $\pm$ 1.764                                                  & 1.753 $\pm$ 0.758    &   6.367 & 3.319                                          \\
                       &  & Dual       & 2.218 $\pm$ 1.837                                                  & 1.739 $\pm$ 0.730    & 6.468 & 3.223                                             \\ \cmidrule(lr){2-7}
& \multirow{3}{*}{Our Method}  & Left only  & 1.861 $\pm$ 1.417  & 1.589 $\pm$ 0.801 & 5.608 & 3.292 \\
                          &                         & Right only & 2.153 $\pm$ 1.431  & 1.569 $\pm$ 0.821 & 5.979 & 3.211 \\
                          &                         & Dual       & \textbf{1.975 $\pm$ 1.427}  & 1.565 $\pm$ 0.815 & \textbf{5.718} & 3.248                                        \\ \bottomrule
\end{tabular}
}
\end{table*}

\subsection{Comparative Performance Analysis}
\label{subsec:comp_anal}
To comprehensively evaluate the proposed method, we conducted three sets of comparisons: (1) against an AprilTag-based pose estimation baseline, (2) against a SOTA monocular visual odometry pipeline, and (3) through an ablation study on the visual servoing module.

For the AprilTag baseline, the pattern shown in Fig.~\ref{fig:exp_apt} was used to compute the world-to-camera transformation via the Perspective-n-Point (PnP) algorithm~\cite{Fischler1981ransac}, based on detected AprilTag corner pixels. Given the known probe tip-to-camera transformation, the probe pose in the world frame was obtained, and pose estimation errors were calculated (Table~\ref{tab:traj_err}). Our proposed method achieved comparable rotational precision to AprilTag, as camera observations are inherently sensitive to rotation. However, by employing visual servoing, our approach significantly reduced translational errors for both the U-shaped ($t$-test, $p=1.072\times10^{-7}<0.05$) and spiral ($p=9.332\times10^{-6}<0.05$) trajectories.

{To benchmark against a high-performing visual odometry method with similar sensing requirements, we evaluated monocular ORB-SLAM3~\cite{ORB-SLAM} without IMU input. In this configuration, ORB-SLAM3 lacks absolute scale observability and therefore required post-hoc alignment with the ground-truth trajectory. Within our constrained workspace, it exhibited noticeable drift and strong sensitivity to calibration errors. Even after optimizing for scale and spatial transformation, its translational and rotational errors remained considerably higher than those of our method (see Table~\ref{tab:traj_err}); for instance, the translational error differences were highly significant on both the U-shaped ($p=3.190\times10^{-12} < 0.05$) and spiral ($p=7.819\times10^{-14} < 0.05$) trajectories.}

Beyond baseline comparisons, we assessed the contribution of the visual servoing component by comparing results with and without this module. As shown in Table~\ref{tab:traj_err}, incorporating visual servoing consistently improves pose estimation accuracy across all camera settings, particularly for translation. $t$-test analysis confirms that these improvements are statistically significant for both the U-shaped ($p=0.038 < 0.05$) and spiral ($p=0.036 < 0.05$) trajectories.

{While our method achieves lower mean errors than the baselines, the standard deviations in Table~\ref{tab:traj_err} remain non-negligible. This variability is primarily attributed to (1) camera calibration imperfections, including lens distortion that is difficult to simulate; (2) minor workspace curvature or warping that impacts image registration; (3) camera noise and motion blur, which can reduce registration accuracy; and (4) residual errors in the Sim2Real calibration process. Further improvements in calibration procedures and workspace setup are expected to reduce this variability in future work.}

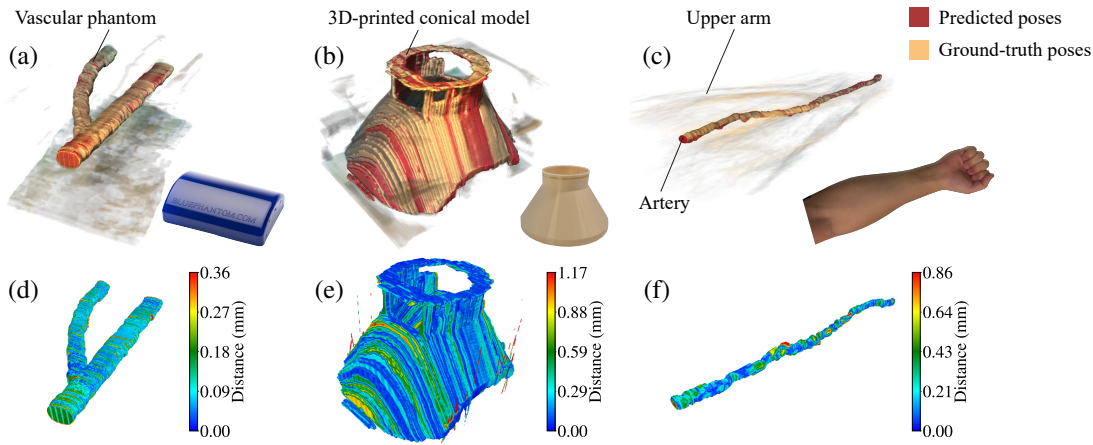
\begin{figure*}[htp]
\centering
\begin{tikzpicture}
    \node[anchor=south west, inner sep=0] (image) at (0,0) 
        {\includegraphics[width=0.8\textwidth]{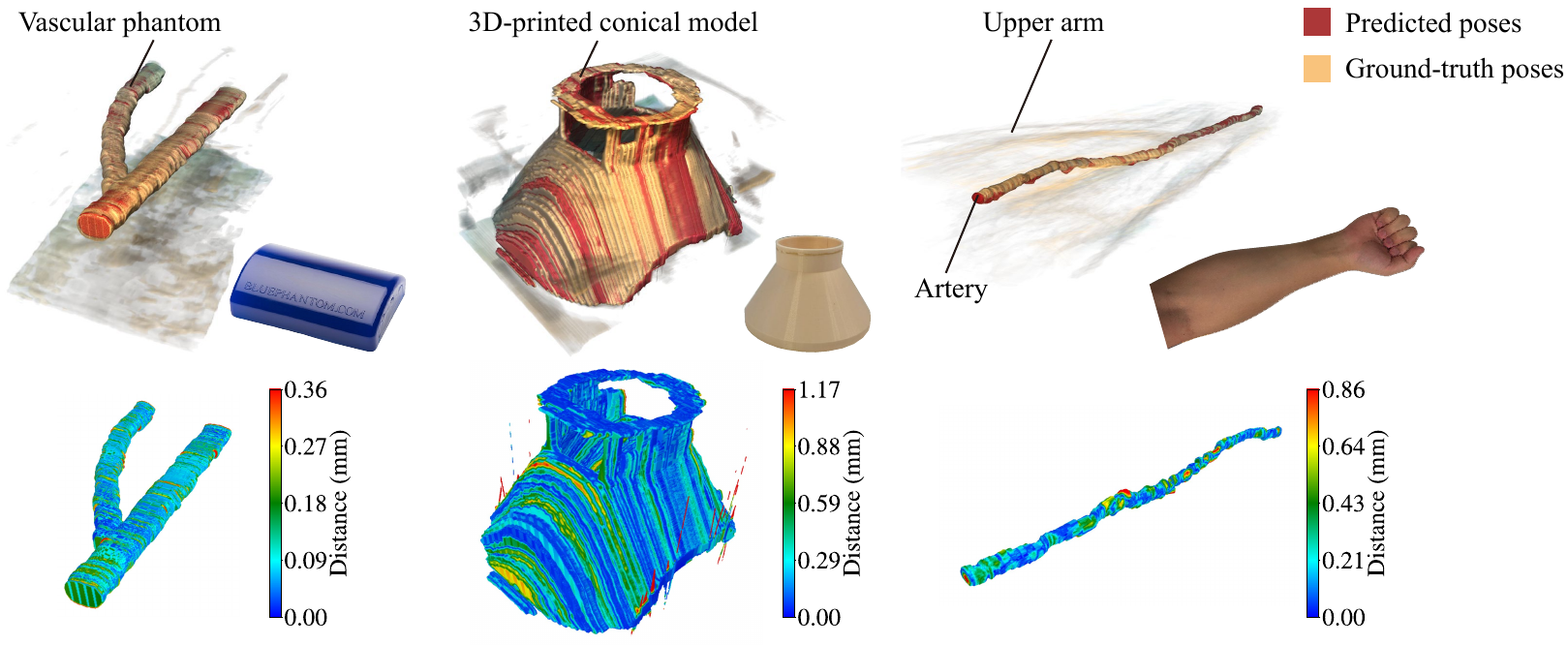}};
    \begin{scope}[x={(image.south east)},y={(image.north west)}]
        \node[black, anchor=north] at (0.02, 0.92) {(a)};
        \node[black, anchor=north] at (0.3, 0.92) {(b)};
        \node[black, anchor=north] at (0.6, 0.92) {(c)};
        \node[black, anchor=north] at (0.02, 0.4) {(d)};
        \node[black, anchor=north] at (0.3, 0.4) {(e)};
        \node[black, anchor=north] at (0.6, 0.4) {(f)};
    \end{scope}
\end{tikzpicture}
\caption{3D reconstruction results from real US scans: (a) Vascular phantom, (b) 3D-printed conical model, and (c) human arm artery. The bottom-right insets in (a)-(c) show the actual scanned objects. In the reconstructions, red regions indicate predicted probe poses and yellow regions represent ground-truth poses. (d)-(f) show colormaps of deviation distances between the predicted and reference reconstructions for (a)-(c), respectively.}
\label{fig:exp_scan}
\end{figure*}

\subsection{{Computational Efficiency Evaluation}}
\label{subsec:comp_cost}
{To further demonstrate the time efficiency of the proposed probe pose estimation method, we compute the time cost for each probe location. All reported time below is computed based on a personal laptop equipped with an Intel i7-12700K CPU and 64~GB RAM, running in Ubuntu~20.04. For individual probe locations, the two camera observations are streamed to the workstation through ROS in real-time.}

{In this study, the image restoration process takes approximately $55.3 \pm 3.4~\mathrm{ms}$ in total for both camera images. This is a one-time operation performed prior to the servoing procedure. The restored images then serve as the target reference for visual servoing. During each iteration of the servoing loop, virtual observations are captured within the simulation environment, the probe motion velocity (including both direction and magnitude) is updated, and the probe is moved to a new pose accordingly. This sequential process takes approximately $88.7 \pm 2.2~\mathrm{ms}$ per iteration. To achieve the desired accuracy, an average of $6.2 \pm 1.5$ iterations are required, with the observed minimum and maximum iteration counts being 5 and 13 in our experiments, respectively. Thus, the total time to compute the probe pose at a single location is approximately $600~\mathrm{ms}$ per camera. It is important to note that real-time pose estimation has not yet been achieved in this study. However, parallelizing the processing of each camera image could further improve efficiency. Future work will focus on algorithmic optimization and accelerating simulation to enable real-time or near-real-time 3D reconstruction for practical applications.}

\subsection{Freehand US Scans and Reconstructions}
\label{subsec:real_scan}
To assess the practicality of our proposed method, we conduct freehand US scans on multiple objects: a vascular phantom (Blue Phantom, CAE, FL, USA), a 3D-printed conical model, and a human arm (live subject). {In these experiments, the Franka robotic arm was set to free-drive mode and guided by the experimenter, simulating a real-world freehand scanning workflow.} {During each scan, we simultaneously recorded US images, dual-camera views, and ground-truth probe poses from the robotic manipulator. All data streams were synchronized using ROS-based message filtering.} {Throughout the experiments, we carefully considered the effects of patient motion and probe-induced tissue deformation on reconstruction accuracy. Our current workflow assumes the scanned object is stationary, and that the region of interest is sufficiently distant from the probe–tissue interface to minimize deformation. For vascular phantom and human arm scans, probe contact was carefully controlled to avoid excessive compression. The 3D-printed conical model was scanned in a water bath, providing acoustic coupling and eliminating deformation concerns.}

US probe poses were estimated from the recorded dual-camera images using our proposed method (see Section~\ref{sec:method}), and 3D reconstructions were generated by integrating these estimated poses with the corresponding US images. Because raw US images often contain substantial noise and extraneous content, we manually annotated the target structures, such as the artery in human arm scans, as its position is relatively stable and well-defined. The annotated images were saved in binary format to enhance reconstruction clarity. The final 3D reconstructions were computed using the annotated images and estimated tracking data within the ImFusion framework\footnote{https://www.imfusion.com/}, as illustrated in Fig.~\ref{fig:exp_scan}. The resulting 3D volumes are well preserved, and reconstructions based on our predicted probe poses closely match those generated using ground-truth robotic poses. {To quantitatively assess 3D reconstruction performance, we report four standard evaluation metrics, which are summarized in Table~\ref{tab:volume}:
\begin{itemize}
\item \textbf{Hausdorff Distance ($d_H$):} Measures the maximum distance from a point in one set ($A$) to its closest point in the other ($B$).
\begin{equation}
d_H(A, B) = \max \Big\{ \sup_{a \in A} \inf_{b \in B} \|a - b\|,\,\, \sup_{b \in B} \inf_{a \in A} \|b - a\| \Big\}
\end{equation}
\item \textbf{Chamfer Distance ($d_C$):} Computes the average distance between all points in one set ($A$) and their nearest neighbors in the other ($B$).
\begin{equation}
d_C(A, B) = \frac{1}{|A|} \sum_{a \in A} \min_{b \in B} | a - b | +
\frac{1}{|B|} \sum_{b \in B} \min_{a \in A} | b - a |
\end{equation}
\item \textbf{Dice Coefficient ($Dice$):} Quantifies the volumetric overlap between two sets ($A$ and $B$) by calculating twice the size of their intersection divided by the sum of their sizes.
\begin{equation}
Dice(A, B) = 2|V_A \cap V_B| / \left(|V_A| + |V_B|\right)
\end{equation}
\item \textbf{Jaccard Index ($J$):} Measures the volumetric ratio of intersection to union between two sets $A$ and $B$:
\begin{equation}
J(A, B) = |V_A \cap V_B| / |V_A \cup V_B|
\end{equation}
\end{itemize}
}

\begin{table}[htp]
\centering
\caption{{Evaluation Metrics for 3D Reconstruction Results}}
\label{tab:volume}
\resizebox{0.45\textwidth}{!}{
\begin{tabular}{llccc} 
\toprule
\textbf{Type} & \textbf{Metric} & {\textbf{Phantom}} & {\textbf{Model}} & {\textbf{Artery}} \\
\midrule
\multirow{2}{*}{\shortstack{Distance\\Metrics}}
  & Hausdorff (mm) $\downarrow$ & 0.359 & 1.171 & 0.858 \\
  & Chamfer (mm) $\downarrow$   & 0.219 & 0.514 & 0.457 \\
\midrule
\multirow{2}{*}{\shortstack{Volume\\Metrics}}
  & Dice Coefficient $\uparrow$ & 0.962 & 0.838 & 0.812 \\
  & Jaccard Index $\uparrow$ & 0.926 & 0.721 & 0.683 \\
\bottomrule
\end{tabular}
}
\end{table}

As shown in Table~\ref{tab:volume}, our method achieves high Dice coefficients and Jaccard indices across all scan targets, demonstrating strong volumetric agreement. The low Hausdorff and Chamfer distances further indicate accurate reconstruction, especially for the vascular phantom and arm artery, while slightly higher values for the conical model reflect its increased geometric complexity. Notably, these errors appear smaller than those reported in Table \ref{tab:traj_err}, likely because the real scanning trajectories in Table \ref{tab:volume} follow mostly unidirectional paths with limited directional and rotational changes, resulting in simpler pose estimation scenarios.

\section{Discussions}
\label{sec:discussion}
In this work, we propose a cost-effective, robust, and precise solution for freehand 3D US reconstruction. Instead of relying on noisy US images for probe pose estimation, we mount two lightweight monocular cameras on the probe to capture distinct visual changes, enabling accurate pose estimation. Compared to US image-based solutions, which suffer from error accumulation over extended scan distances, our method estimates a global, tissue-independent probe pose, offering superior accuracy and stability for long US scan sweeps. Furthermore, compared to the AprilTag-based localization method, our approach enhances translational accuracy. This improvement stems from our closed-loop image matching mechanism, which allows for precise image alignment. Additionally, we introduce an image restoration technique to recover details obscured by occlusion or lighting variations. This reduces the discrepancy between simulated and real-world images. With these merits, our method proves to be a precise, robust, and practical solution for US probe pose estimation and freehand 3D US reconstruction.

{Our method can benefit clinical workflows by enabling clearer anatomical visualization for more reliable diagnosis, reducing the dependence on expensive tracking equipment, and allowing greater accessibility in point-of-care and resource-limited settings. The system also supports medical training and skill assessment by providing immediate, reproducible feedback on probe positioning and 3D reconstruction results. These features can help facilitate broader clinical adoption and integration of advanced US techniques.}

Our method also has certain limitations. First, it cannot tolerate a fully occluded camera FoV. If the scanned object is too large and fully obstructs the cameras' FoV, adjustments are necessary, such as repositioning the workspace pattern to the side of the object or modifying the camera mount. {The camera arrangement in our system is empirically selected to increase workspace observation coverage and reduce the effect of potential obstruction. It is worth noting that the primary goal is to provide complementary visual coverage rather than enable stereo triangulation in this study. The current setup can be extended to have more than two cameras to further improve robustness in complex scenarios.} {Second, we currently rely on the Franka Emika Panda robotic arm to provide ground-truth probe poses for both hand-eye calibration and Sim2Real calibration.} Therefore, the overall accuracy evaluation is limited by the positioning precision of the robotic system, which in our case is approximately $0.5~\mathrm{mm}$. Future studies may incorporate a laser tracker to enable more precise localization assessment. {A third limitation is that real-time pose estimation has not yet been achieved in this study. Parallelizing camera image processing and optimizing the algorithm could further improve computational efficiency.}


\section{Conclusions and Future Work}
\label{sec:conclusion}

This work presents a cost-effective probe pose estimation algorithm {designed for freehand US reconstruction}, utilizing two lightweight cameras. These cameras capture visual feedback from a planar workspace featuring a textured pattern. To explicitly address practical challenges such as occlusions caused by patients or operators and lighting variations, we introduce an image restoration technique. For accurate pose estimation, a sim-in-the-loop framework is developed. This framework simulates the real-world system setup and iteratively minimizes the pose error between simulated and real-world observations using a PBVS controller. This approach significantly improves translational estimation accuracy by precisely aligning simulated and real camera observations. Experimental evaluations across various spatial trajectories demonstrate that our method achieves high accuracy in both positional and rotational estimations. Freehand US scans and 3D reconstructions performed on a soft vascular phantom, a 3D-printed conical model, and a human arm yielded Hausdorff distances to the reference reconstructions of 0.359 mm, 1.171 mm, and 0.858 mm, respectively. These results validate the feasibility of our approach for practical applications. Future work will include prospective clinical trials to evaluate system usability and diagnostic accuracy in real clinical settings. We will also explore compatibility with various probe types and advanced camera configurations to further improve robustness.

\appendices
\section{{AprilTag-Based Baseline Setup}}
\begin{figure}[htp]
\centering
\includegraphics[width=0.42\textwidth]{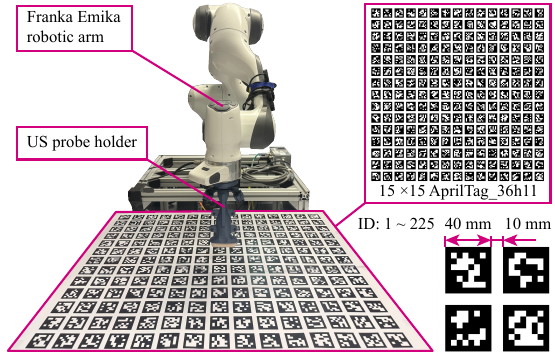}
\caption{Experimental setup using a dense AprilTag pattern for quantitative baseline comparison. AprilTag results are reported as a baseline reference; all results of our proposed method use the workspace pattern.}
\label{fig:exp_apt}
\end{figure}

AprilTag is used as a baseline to prove robust and highly accurate fiducial marker detection. In this study, each AprilTag measures $40\times40~\mathrm{mm}$ with a $10~\mathrm{mm}$ separation in both X and Y directions, forming a $15 \times 15$ grid (see Fig.~\ref{fig:exp_apt}). The use of dense AprilTag markers can significantly enhance the pose estimation performance compared to sparse markers because the dense version will enforce both local constraints when detecting individual markers, and all detected markers will guarantee the global constraints.

\bibliographystyle{IEEEtran}
\bibliography{reference}
\vspace{-10 mm}

\end{document}